%% file: acl2023.tex
\pdfoutput=1
\documentclass[11pt]{article}

\usepackage[]{ACL2023}

\usepackage{times}
\usepackage{latexsym}
\usepackage{booktabs}
\usepackage{eqparbox}
\usepackage{arydshln}
\usepackage{subcaption}
\usepackage{rotating,csquotes}
\usepackage{xcolor}
\usepackage{multirow}
\usepackage{framed}
\usepackage{booktabs}
\usepackage{amsmath}
\usepackage{amssymb}
\usepackage{mathtools}
\usepackage{upgreek}
\usepackage{soul}
\usepackage{amsfonts}
\usepackage[most]{tcolorbox}
\usepackage{tabularx}

\usepackage[T1]{fontenc}
\usepackage[utf8]{inputenc}

\usepackage{microtype}

\usepackage{inconsolata}

\title{A User-Centric Multi-Intent Benchmark for\\Evaluating Large Language Models}
\author{Jiayin Wang$^1$, Fengran Mo$^2$, Weizhi Ma$^{3*}$, Peijie Sun$^1$, Min Zhang$^{1*}$, Jian-Yun Nie$^2$\\
$^1$Department of Computer Science and Technology, Tsinghua University, Beijing, China. \quad \\
$^2$Université de Montréal, Québec, Canada. \quad \\
$^3$Institute for AI Industry Research, Tsinghua University, Beijing, China.\quad \\
\texttt{JiayinWangTHU@gmail.com, mawz@tsinghua.edu.cn, z-m@tsinghua.edu.cn}}

\input{definitions}

\begin{document}
\maketitle
\begin{abstract}

%With the rapid development of 
% Large Language Models (LLMs) are essential tools that users may use in various scenarios for distinct intents.
Large language models~(LLMs) are essential tools that users employ across various scenarios, so evaluating their performance and guiding users in selecting the suitable service is important.
Although many benchmarks exist, they mainly focus on specific predefined model abilities, such as world knowledge, reasoning, etc. 
Based on these ability scores, it is hard for users to determine which LLM best suits their particular needs.
To address these issues, we propose to evaluate LLMs from a user-centric perspective and design this benchmark to measure their efficacy in satisfying user needs under distinct intents.
Firstly, we collect 1,846 real-world use cases from a user study with 712 participants from 23 countries. This first-hand data helps us understand actual user intents and needs in LLM interactions, forming the \textbf{U}ser \textbf{R}eported \textbf{S}cenarios~(URS) dataset, which is categorized with six types of user intents. 
Secondly, based on this authentic dataset, we benchmark 10 LLM services with GPT-4-as-Judge.
Thirdly, we show that benchmark scores align well with human preference in both real-world experience and pair-wise annotations, achieving Pearson correlations of 0.95 and 0.94, respectively. 
This alignment confirms that the URS dataset and our evaluation method establish an effective user-centric benchmark.
The dataset, code, and process data are available\footnote{\url{https://github.com/Alice1998/URS}}.
% \footnote{\url{https://anonymous.4open.science/r/URS/}}.
% The range of these applications spans from information retrieval to the pursuit of creativity.

% The rapid advancement of large language models (LLMs) is profoundly impacting and transforming the way people work and live. LLMs have demonstrated immense practical potential in various scenarios, such as text generation and logical reasoning, due to their powerful processing capabilities. 
% However, current evaluation methods of LLMs primarily focus on their professional abilities in logical reasoning, mathematics, and coding, treating them as independent intelligence, sometimes neglecting their performance in other dimensions, such as creativity and anthropomorphism. As LLMs are not only standalone entities but also collaborators that assist users in completing various daily or complex tasks, there is a need for a new evaluation aspect that considers their widespread applications in people's daily lives and professional activities. This paper aims to assess the ability of LLMs to satisfy user needs and provide high-quality responses in real-world usage scenarios, such as information retrieval, solving professional problems, text processing, offering advice, generating ideas, and leisure entertainment, from the perspective of user intentions.
\end{abstract}

\input{section/1_introduction}
\input{section/2}

\clearpage

\bibliography{anthology,custom}
\bibliographystyle{acl_natbib}

\appendix
\input{section/appendix}

\end{document}

%% file: definitions.tex
\usepackage{graphicx}
\usepackage{makecell}
\usepackage{multirow} 
\usepackage{diagbox}
\usepackage{tabularx}
\usepackage{amsmath}
\usepackage{pifont}
\usepackage{url}

\usepackage{CJKutf8}
\newcommand{\chinese}[1]{\begin{CJK}{UTF8}{gbsn}#1\end{CJK}}
\usepackage{tcolorbox}
\usepackage{color}

\usepackage{colortbl}

\tcbset{
  aibox/.style={
    width=450pt,
    top=10pt,
    colback=white,
    colframe=black,
    colbacktitle=black,
  }
}
\newtcolorbox{AIbox}[2][]{aibox,title=#2,#1}

%% file: section/1_introduction.tex
\section{Introduction}
\label{sec:intro}
% Points:

% 1 seven intents:
% diverse real-world scenario

% 2 user-reported: user cared cases, general without selection bias towards particular applications.

% 3 evaluations

% contribution:
% 1)
% 2)
% 3)

Large language models (LLMs) are rapidly developing and gradually changing the way people interact with computing systems and permeating diverse facets of work and daily life~\cite{wang2023enabling}.
Although LLMs show powerful capacities for completing various tasks~\cite{chang2023survey}, it is essential to understand how they satisfy users' actual intents and needs~\cite{elangovan2024considers}. 
% An intuitive way to understand the satisfaction in interactions with LLMs is to conduct evaluations. 
According to related studies, many attempts have been made to evaluate LLMs, which may be divided into following two groups.
%As LLMs become increasingly integrated into people's lives, it is essential to evaluate their efficacy from the perspective of satisfying user intents and needs.
% These models possess remarkable capabilities in processing and generating natural language, enabling them to excel in tasks such as text generation and logical reasoning. 
% As LLMs continue to advance and find widespread application in both personal and professional contexts, it becomes increasingly important to evaluate their performance from a user-centric perspective.
\begin{figure}
    \centering
    \includegraphics[width=\linewidth]{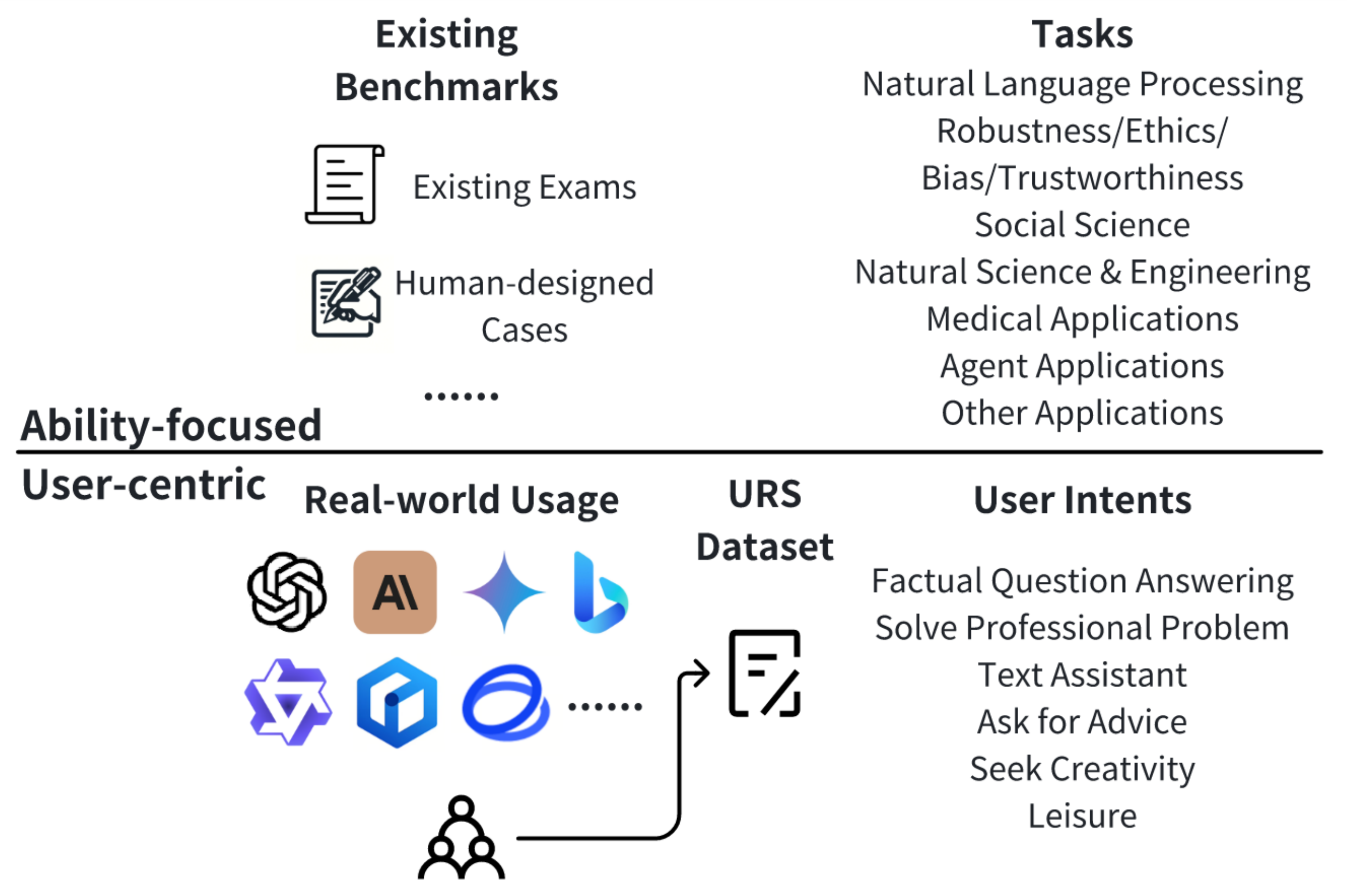}
    \caption{Existing benchmarks are mainly model ability-focused and categorized by tasks~\cite{chang2023survey}. We benchmark LLMs on \textbf{U}ser \textbf{R}eported \textbf{S}cenarios~(URS), which are user-centric, intent-driven, multi-cultural, and multi-LLM usage cases involved.}
    \label{fig:main}
\vspace{-2ex}
\end{figure}

\begin{table*}[t]
    \centering
    \begin{tabular}{lrlll}
    \toprule
    \textbf{Benchmark} & \textbf{Cases} & \textbf{Language} & \textbf{Source} &  \textbf{Category}  \\ 
    \midrule
    % MMLU, etc. &  & & core knowledge & & &  \\
    % Flan, etc. & & & instruction-following & & \\
    % CoQA, etc.& & & conversational & & &  \\
    AlpacaEval & 805 & English & existing datasets and synthetic data & 5 sources\\
    MT-Bench-101 & 1,388 & English & synthetic data & 13 tasks\\
    TencentLLMEval & 3,000& English, Chinese & human designed and written tasks  & 7 areas \\
    MT Bench & 80 & English  & human written questions  & 8 tasks  \\
    AlignBench & 683 & Chinese & selected questions in ChatGLM logs &   8 tasks \\
    WildBench & 1,024 & English & selected questions in ChatGPT logs &  12 tasks \\
    \midrule
    URS (Ours) & 1,846 & English, Chinese & self-reported logs with 15 LLMs & 6 intents \\
    \bottomrule
    \end{tabular}
    \caption{Comparison between user-related LLM benchmarks with our URS benchmark. The compared studies are AlpacaEval~\cite{alpaca_eval}, 
    MT Bench-101~\cite{bai2024mt}, TencentLLMEval~\cite{xie2023tencentllmeval},
    MT Bench~\cite{zheng2024judging}, AlignBench~\cite{liu2023alignbench} and
    WildBench~\cite{wildbench2024}.}
    \label{table:contribution}
\vspace{-2ex}
\end{table*}

The first group of work~\cite{hendrycks2020measuring,clark2018think,zellers2019hellaswag} considers LLMs as task-completion models and mainly employs close-domain assessments to measure their various task-solving abilities, such as world knowledge, mathematics ability, coding skill, reasoning, planning, agent task, etc.~\cite{chang2023survey}.
However, the existing attempts focused on predefined tasks, without considering the roles that humans would like LLMs to play %it neglects the collaborative and supportive roles LLMs play alongside humans 
in real-world scenarios. For example, while a task was evaluated, it was unclear if the task was important to humans and frequently required by users in real applications.
% and cannot evaluate whether they can truly satisfy users' various intents.

The second group evaluates whether LLMs' responses align with human preferences~\cite{wang2023aligning}.
%As shown in 
Table~\ref{table:contribution} summarizes the user-related LLM evaluation benchmarks. Their evaluation datasets include synthetic data~\cite{alpaca_eval}, human-written data~\cite{zheng2024judging}, or selected user logs from certain LLM services~\cite{wildbench2024}. % are provided for evaluation.
We can observe that these benchmarks may be limited in terms of data quality, quantity, or focus on a single resource. They do not reflect the true distribution of intended utilization of LLMs by real users and their effectiveness for these intents. %However, the first two benchmarks are not enough in quality or quantity, while the remainings faces the problem of a single source that may deviate from real-world distributions as users may have limited access to certain LLMs and deliberately select LLM services based on intents.
%Detailed discussions between our work and theirs are provided in Section~\ref{section:related_work_benchmark}.
% which is not enough in terms of quality, quantity and distribution alignment with real-world scenarios. 

Besides, the categorization in current benchmarks is mainly focused on single-step %model completed 
tasks~\cite{zhao2023inthe, liu2023alignbench}.
However, end users might be more interested in knowing the ability of LLMs to satisfy their intended needs, rather than  %not be aware of 
LLMs' performance on predefined individual tasks. % and abilities, but interested in whether their intents could be satisfied by current services or not.
In particular, a user intent %The individual user intent 
might involve multiple model abilities with a series of tasks.
For example, when a user is asking for advice about travel plans, this might involve model abilities of planning, personalization, and calling APIs. 
Consequently, the categorization of model-side abilities complicates the assessment of LLM performance in user-side scenarios, making it challenging for users to select the suitable service tailored to their specific needs.

This paper addresses the above issues in LLM evaluations by developing a user-centric benchmark to examine whether LLMs' behaviors satisfy user needs in real-world scenarios.
% For data construction, we collect large scale real-world use cases from the user side, aiming to build a dataset corresponding to the true usage of LLMs. 
% Besides, this approach mitigate bias towards certain LLMs, which 
% For evaluation, we design an intent-aware auto evaluation method, which is examined to have alignment with human preferences.
% as the public release of user logs needs users' approval to avoid privacy concerns~\cite{feldhus2023interrolang,bolotova2022non}. 
% assessing LLMs' ability to deliver clear and concise responses that align with real user needs under both objective and subjective user intents. 
% Note that we are a third-party benchmark designer, without being associated with a specific LLM service.
% Our benchmark is designed on a third-party evaluation, without being associated 
% %and no specific association 
% with a specific LLM service. %services are required.
As shown in Figure~\ref{fig:main}, our work is different with the existing ability-focused benchmarks and highlighted with the three unique characteristics:
% Our differences with the existing ability-focused benchmark are shown in Figure~\ref{fig:main}, which highlights the three unique characteristics of our work: % are in three parts:

\noindent \textbf{User-centric.} 
We benchmark LLMs from the user perspective in both dataset construction and evaluation designs.
% Our URS dataset contains first-hand usage scenarios across different LLM services from real users, ensuring the freshness and diversity. 
Our constructed \textbf{U}ser \textbf{R}eported \textbf{S}cenarios (URS) benchmark encompasses 1,846 authentic interactions~(1,014 in English and 832 in Chinese) across 15 diverse LLM services, contributed by 712 users from 23 countries, each vetted through third-party manual quality checks. This dataset, reflecting the real-world usage of LLMs, is publicly available for research with user consent.
Based on this first-hand data, we design evaluations to measure LLMs' efficacy in satisfying real user needs.
The benchmark results align well with human preference in both real-world user experience across intents and pairwise comparison between LLMs.
% The real-world scenarios often simple in articulation, embody complexity in their resolution and span an expansive range of situations.
% The practical cases in this URS dataset enable us to provide a more accurate gauge of LLMs' efficacy as cooperative intelligent services in satisfying user needs.
% by evaluating LLMs against such practical cases. 
% This user-centric approach encourages LLMs to closely collaborate with humans as their capabilities continue to improve.
This user-centric approach promotes deeper collaboration between LLMs and humans as their capabilities continue to improve.

% Our benchmark focuses on evaluating LLM from users' perspective.
% The dataset is derived from actual use cases self-reported by users across 20 distinct LLM platforms.
% Real-world user problems are often simple in description and complex in task, and cover a wide variety of use cases. Evaluating LLMs on such problems can more realistically measure the performance of each LLM as a collaborative tool to meet the user's needs, and can help the user to choose the appropriate LLM service according to his/her needs. At the same time, from the perspective of model development, promoting LLM in understanding the user's tasks in real scenarios can make LLM align with user preference while improving its basic capabilities, which promotes the continuous development of LLM.

% \todo{exmaple, one intent - task, }
\noindent \textbf{Intent-driven.} 
Different from the task-specific evaluations, %ability-focused studies, 
our benchmark is categorized by user self-selected intents, including factual question answering, professional problem solving, text assisting, asking for advice, seeking creativity, and leisure.
% , and usage through API.
Such categorization provides the potential utilization of LLMs by real users and is validated through the user study. %more intuitive framework by considering that the users may anticipate different types of feedback depending on their specific intents. 
% In real-world scenarios, the user intents often demand a blend of model abilities to address a series of tasks. 
Based on the divided intents, the users who lack specific knowledge about the model capabilities can simplify their choice of proper service. For researchers and developers, this categorization helps to provide a more precise and targeted evaluation of how well LLMs meet user needs in diverse scenarios as one simple user intent may demand a blend of model abilities to address a series of tasks~\cite{bolotova2022non}. 

%Acknowledging that users may anticipate different types of feedback depending on their specific intents, and that real-world scenarios often demand a blend of model abilities to address a series of tasks, this categorization provides a more intuitive framework. 
%It prioritizes user intents over traditional taxonomy of model capabilities, allowing a more precise and targeted evaluation of how well LLMs meet user needs in each scenario. This approach also simplifies the user's choice of proper service based on a specific intent, as users may lack familiarity with the underlying model capabilities, and it is fairly straightforward to determine their own intents. 
% It prioritizes user intents over traditional taxonomy of model capabilities, enhancing the relevance and accessibility of LLM services.
% This classification allows users, who may not be familiar with model capabilities, to easily select LLM services that are optimally tailored to their specific needs. This intent-based taxonomy shifts the focus from the traditional model ability framework to a more user-centric approach.

\noindent \textbf{Multi-cultural.} The URS data is contributed by users from 23 countries in Asia, Europe, North America, Oceania, South America, and Africa. Their reported cases cover multiple cultural backgrounds, such as nations' traditional festivals, local points of interest, and pop culture across the globe.
This brings diversity in LLM evaluations, which is not included in simple translations of English-dominant content~\cite{hershcovich2022challenges}.

In this study, our contributions are threefold:

(1) We collect 1,846 authentic cases from 712 global users to form the multi-intent, multi-cultural \textbf{U}ser \textbf{R}eported \textbf{S}cenario dataset.

(2) Based on this URS dataset, we propose an intent-aware evaluation method to benchmark the efficacy of 10 LLM services in satisfying user needs in diverse real-world scenarios.

% (3) Experiment results validate the effectiveness of this user-centric benchmark: Intent-wise benchmark scores correlate with user-reported experience in LLM interactions (r = 0.95). Furthermore, LLM-wise benchmark orders align with human pairwise comparisons (r = 0.94).

(3) Experiments demonstrate that benchmark results align closely with human preferences, as evidenced by Pearson correlations of 0.95 and 0.94 with real-world user experiences and pairwise annotations. 
% These results validate the effectiveness of this user-centric benchmark approach.
These results validate that our automated evaluation method and URS dataset establish a new and effective user-centric benchmark.

%% file: section/2.tex
\section{Related Work}

\subsection{LLM Benchmark}
\label{section:related_work_benchmark}
The existing benchmarks are mainly divided into two categories as illustrated in Figure~\ref{fig:main}. They are model ability-focused and user-centric.

\noindent \textbf{Ability-focused benchmarks.}
Considering the broad capabilities of Large Language Models (LLMs), the evaluations span a diverse range of domains, typically categorized into 
Knowledge: MMLU~\cite{hendrycks2020measuring}, CMMLU~\cite{li2023cmmlu}, and C-Eval~\cite{huang2024c}; Mathematical abilities: GSM8k~\cite{cobbe2021gsm8k}, MATH~\cite{hendrycks2021measuring}, and OlympiadBench~\cite{he2024olympiadbench};
Coding skills: HumanEval~\cite{chen2021evaluating}, MBPP~\cite{austin2021program}, or SWEBench~\cite{jimenez2023swe}; Reasoning: BBH~\cite{suzgun2022challenging, srivastava2023beyond}, ARC~\cite{clark2018think}, and HellaSwag~\cite{zellers2019hellaswag};
Agent tasks: AgentBench~\cite{liu2023agentbench} and ToolBench~\cite{qin2023toolllm}.
Each category aims to measure specific abilities expected from LLMs. 

\noindent \textbf{User-Centric benchmarks.}
Unlike benchmarks that focus on model capabilities, 
% this work aim to measure LLMs' efficacy in satisfying real-world user needs.
a few emphasize user experience with chat-based LLMs. 
Table~\ref{table:contribution} details related user-centric benchmarks. However, we notice several problems with these benchmarks. % for the intended utilization of LLMs by real users.
AlpacaEval~\cite{alpaca_eval}, MTBench~\cite{zheng2024judging}, and MT-Bench-101~\cite{bai2024mt} are limited in data quality or quantity.
TencentLLMEval~\cite{xie2023tencentllmeval} used tasks that were designed by hand and written by 50 annotators, an approach that may lead to deviation from real-world scenarios. 
AlignBench~\cite{liu2023alignbench} and WildBench~\cite{wildbench2024} are constructed based on real-world use cases from logs of one specific LLM, which may introduce bias since the users may have access limitations to certain LLMs and use LLM services selectively depending on their needs. Furthermore, the existing user-centric benchmarks mainly do not consider multilingual %they lack the linguistic 
and multi-cultural aspects, which are covered in our dataset.
% In contrast, WildBench~\cite{wildbench2024} gathered user questions from WildChat~\cite{zhao2023inthe}. However, these questions predominantly originate from interactions with ChatGPT, potentially leading to a biased distribution. Furthermore, they lack the linguistic diversity in our dataset.
% Different from them as shown in table~\ref{table:contribution}, our studies focus on evaluating LLM on user self-reported scenarios, which contain both English and Chinese background and are divided by user-reported intents.

% Table~\ref{table:contribution} summarizes the distinction between our benchmark and most related studies, which are all categorized by model side task-solving abilities.
% Among them, AlignBench~\cite{liu2023alignbench} and WildBench~\cite{wildbench2024} are constructed based on real-world use cases from the user logs, but they only report the selected cases and the sizes are smaller. Besides, they only access one specific LLM, which may introduce bias since the users may use LLM services selectively depending on their needs and access ease.
% Based on the URS dataset, we can provide a more authentic picture of user satisfaction in interacting with LLMs in real scenarios with various intents.

\subsection{Evaluation Methods}
% \todo{类型 好处}
Besides dataset constructions,
evaluation methods are crucial in developing benchmarks. To simplify evaluation processes, many studies utilize multiple-choice questions as criterion~\cite{hendrycks2020measuring, li2023cmmlu, huang2024c}. Others employ post-processing of generated content and apply rules or deterministic tests to assess performance~\cite{cobbe2021gsm8k, chen2021evaluating}. With the recent advancements in LLM techniques, an increasing number of approaches use a more capable AI model to evaluate answers~\cite{he2024ultraeval,alpaca_eval}, enriching benchmark types by allowing a more diverse format of test questions.
AlpacaEval~\cite{alpaca_eval}, AlignBench~\cite{liu2023alignbench} adopt GPT-4 as evaluator. PRE~\cite{chu2024pre} shows that GPT-4 performs well in pairwise evaluations and ~\citet{huang2024empirical} shows that GPT4 outperforms fine-tuned judge models in generalizability and fairness. 
% This LLM-based evaluation method enriches benchmarks by allowing a more diverse format of test questions. 
In our benchmark, we adopt previous settings for open-domain evaluation, using LLM-based pairwise evaluation.

\subsection{User Intent}
% For the categorization of LLM benchmarks, 
% a line of work focuses on model completed tasks~\cite{wildbench2024, zheng2024judging}
% , such as information seeking, creative writing, coding \& debugging, reasoning, editing, Math, planning, brainstorming, role-playing, advice seeking, data analysis.
% This taxonomy is closely related to the single-step capabilities of LLM.
From a user-centric view, we should first confirm user intent and evaluate service responses correspondingly.
% To understand user satisfaction, it is important first to determine their intents.
Related ideas try to understand user intent in interacting with LLMs.
\citet{shah2023using} provides a user intent taxonomy based on New Bing, which includes information retrieval, problem solving, learning, content creation, and leisure. \citet{bodonhelyi2024user} details user intent with ChatGPT with informational, problem solving, creative, educational, personal interaction, technical and professional, transactional, ethical, and philosophical intents.
%Based on these previous ideas about user intent taxonomy, we design our dataset categorization and illustrate it in Section~\ref{section:user_intent}.
Inspired by these studies, we design a user intent taxonomy accordingly, which is further verified by 712 user study participants.

% based on our poll and our benchmark accordingly. %for our proposed dataset.

% They provide taxonomy of 
% another line of categorization is understanding user intents with conversational system~\cite{shah2023using,bodonhelyi2024user}. This taxonomy includes factual question answering, solving professional problems, text assistance, asking for advice, seeking creativity, leisure, and usage through API.
% In our work, we categorize the dataset based on user intents, instead of model abilities.

\section{User Reported Scenarios Dataset}
% To better understand the real needs and utilization of LLMs by users, we conducted a survey to collect the recent use cases of LLMs. From the survey, we verify the user intent taxonomy.

\subsection{User Intent Taxonomy}
\label{section:user_intent}
% \section{User Reported Scenarios Dataset}
%To collect the first-hand use cases of general LLMs, we design and conduct a user study.
% In the following subsections, we present the design and distribution process of this study.

% \subsection{User Survey}

% \subsection{User Intents}
%To understand user satisfaction, it is important to first understand the intents of users when they interact with LLMs.
User intents represent specific needs or goals when interacting with a service and users may anticipate different types of feedback under different intents~\cite{bolotova2022non}. For example, for factual intents, users may expect brief and accurate answers; for brainstorming needs, users tend to expect rich and innovative answers.
By considering different user intents, we can assess LLMs' fine-grain efficacy in diverse scenarios.
% This benchmark aims to evaluate LLMs' performance as collaboration tools to meet user needs, thus we categorize the dataset according to user intents.
%Based on related ideas,
Therefore, we first define the taxonomy of user intent based on related work~\cite{shah2023using,bolotova2022non}. They are detailed below:
% This taxonomy is further verified in the following user study.
%According to related study~\cite{shah2023using,wang2024understanding}, this work divides user intents into objective intents and subjective intents. The taxonomy design is described below:

% \subsubsection{Objective Intents} 

\noindent  \textbf{Factual QA.}
Fast and direct access to factual information.
For example, search for historical events, scientific facts, or public data.

\noindent  \textbf{Solving Professional Problems.}
Require answers and insight in specialized fields like natural sciences, humanities, or social sciences for problem-solving or learning. It involves domain knowledge, in-depth understanding, and reasoning in professional areas.
For example, solving math, engineering puzzles, and conducting medical diagnoses.

\noindent  \textbf{Text Assistant.}
Need assistance with text-related tasks such as summary, translation, editing or completing content.
For example, replying to an email, writing a report, or polishing a speech.

% \noindent  \textbf{Research Purpose Usage through API.}
% A frequent intent is to ask an LLM to utilize Application Programming Interface (API) instead of exploiting %user interfaces for a more technical exploration of 
% LLM's own abilities to solve the problem.
% For example, one may evaluate LLMs on various tasks~(such as testing their mathematical computing or generating SQL queries to retrieve data from a database), simulating agents, environments, or datasets.
% Exploring through the Application Programming Interface~(API) of LLMs instead of the web chat interfaces for LLM-related research.
% For example, one may call LLM interfaces to benchmark their performances and simulate datasets, environments, and agents with LLM services' API.

% \subsubsection{Subjective Intent} 

\noindent  \textbf{Ask for Advice.}
Look for opinions and suggestions for personal or professional decisions, including planning or counseling.
For example, career development, personal counseling, creating travel plans, or shopping lists.

\noindent  \textbf{Seek Creativity.}
Brainstorming for inspiration and innovative ideas.
For example, advertising ideas, or design inspiration.

\noindent  \textbf{Leisure.}
Engage in or look for recommendations for recreational activities, including books, music, movies, games, and other entertaining activities.

This user intent taxonomy serve as the potential utilization of LLM services by real users and is furthered verified in the following user study.

\begin{figure}[t]
    \centering
    \includegraphics[width=\linewidth]{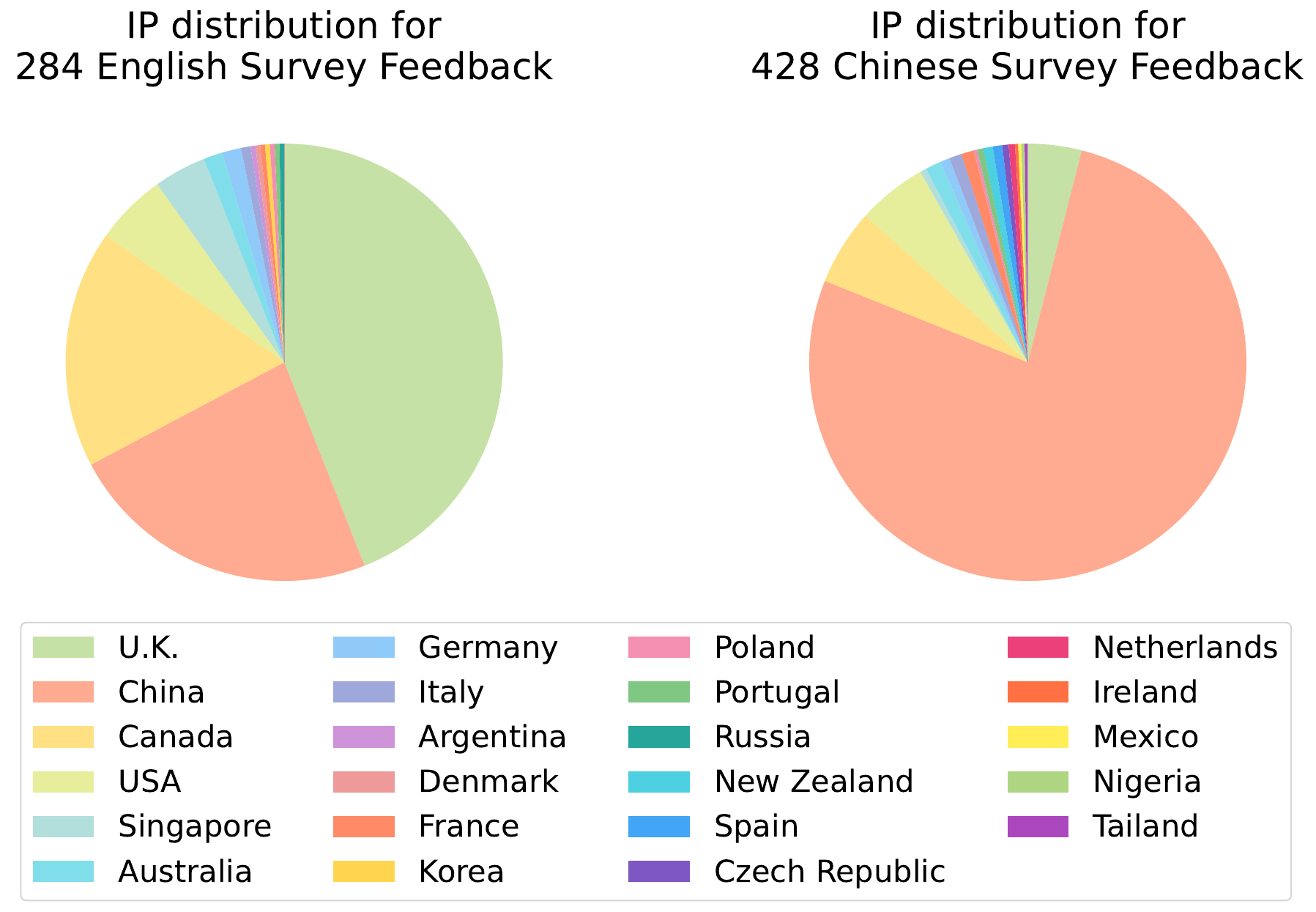}
    \caption{IP Distribution of the 712 participants.}
    \label{fig:IP}
\vspace{-2ex}
\end{figure}

\begin{table}[t]
\begin{tabular}{lrrr}
\toprule

  & \textbf{English} & \textbf{Chinese} & \textbf{Total}  \\    \midrule
Total             & 1,014    & 832     & 1,846\\ \midrule
 % \multicolumn{3}{l}{Objecti ve Intents: 66\%}  & \\ 
Factual QA      & 374     & 199     & 573  \\
Solve Problem   & 180     & 348     & 528  \\
Text Assistant  & 81      & 43      & 124  \\  \midrule
% Use through API             & 14      & 14 & 28\\         \midrule
 % \multicolumn{3}{l}{Subjective Intents: 34\%}  & \\ 
Ask for Advice  & 175     & 94      & 269  \\
Seek Creativity & 121     & 72      & 193  \\
Leisure         & 83      & 76      & 159  \\         
\bottomrule
\end{tabular}
\caption{User Intent Distribution in the URS dataset. }
\label{table:satistic_intents}
\vspace{-2ex}
\end{table}
% \todo{group by continent}

% This intent taxonomy is provided in the user study.

\subsection{Dataset Development}
\label{sec:dataset_dev}
To better understand LLMs' applications in real-world scenarios, we conducted a user study to collect participants' recent interactions with LLM services, following data collection strategies of related studies~\cite{feldhus2023interrolang,bolotova2022non}.
\subsubsection{User Study}
\label{section:user_study}
Participants in the study were requested to provide five to ten real cases with LLMs they have used. Each reported case included the following three components:
the LLM service they queried,
complete conversation, and their usage intent~(either selected from the predefined list or filled in by themselves). The detailed questionnaire is attached in Appendix~\ref{sec:appendix}.

Additionally, the study investigate users' general experience with LLMs, including their satisfaction across different intents, which is used to validate benchmark alignment with human feedback in Section~\ref{section:sec:alignemt}.
We also record the anonymous, voluntarily reported demographic information in the questionnaire to examine whether this survey involves diverse groups of participants.

%The questionnaire also records basic demographic information, including age group and occupation, to help us examine whether it is distributed to diverse groups of participants.
% For demographic information, we record age, occupation, and IP location, where age and occupation can be optionally not reported. 

% \noindent  \textbf{Distribution.}
To capture multi-cultural scenarios, we offer the questionnaire in both English and Chinese and spread it through Prolific\footnote{https://www.prolific.com. A global human-label platform for research and AI.}, X, and WeChat Moments platforms for global enrollment. Crowdworkers are paid according to Prolific recommendations~(9£/hr * average 10 mins = 1.5£).

\begin{table*}[t]
\centering
% \resizebox*{0.99\textwidth}{!}{
\begin{tabular}{p{3.5cm}p{11.5cm}}
\toprule
\textbf{Intent} & \textbf{Example Cases} \\ \midrule
Factual Question \newline Answering & EN: Tell me what bitcoin is \newline CN: \chinese{大雪农历初几} (\textit{When is the Major Snow in the Lunar calendar}) \\ \midrule
Solve Professional \newline Problems & EN: How long does it take to transfer 13.72GB if speed is 10MB/s \newline CN: \chinese{假设你正在爬楼梯。需要n阶你才能到达楼顶。每次你可以爬1或2个台阶。你有多少种不同的方法可以爬到楼顶呢？}  (\textit{Suppose you are climbing a staircase. It takes n steps to reach the top. You can climb 1 or 2 steps at a time. How many different ways can you climb to the top?}) \\ \midrule
Text Assistant & EN: Help me rephrase the document: The NBA's inaugural in-season tournament has concluded with the Los Angeles Lakers beating the Indiana Pacers 123-109 to lift the NBA Cup, with the team's winning players ......\newline CN: \chinese{请你帮我撰写一段给领导2024龙年的拜年微信。}  (\textit{Help me compose a Spring Festival's Eve Wechat text to my leader for the Year of the Dragon 2024.})\\ \midrule
% Usage through API & EN: Don't use any tools! **Directly** tell me the answer (no intermediate processes): 23 + 32 + 35 + 25 + 32 + 34 \newline CN: \chinese{“护肤水已经空瓶啦，肤感不错，保湿效果很好，而且成分安全”的情绪是消极的还是积极的} (\textit{"The skin lotion is empty, feels good, moisturizes well, and ingredients are safe." Is the sentiment negative or positive?}) \\ \midrule
Ask for Advice & EN: How long can vegan blueberry croissants be kept for once they've been baked off from frozen \newline CN:\chinese{在8度天气里，适合穿哪些款式和类型的外套？}  (\textit{What styles and types of jackets are appropriate in 8 degree weather?}) \\ \midrule
Seek Creativity & EN: Explain photosynthesis as if I am 9 years old. \newline CN: \chinese{帮我的猫咪起个好听的英文名字，她的中文名字叫小惰娇，1岁的女娃
} (\textit{Help me give my cat a nice English name. Her Chinese name is Xiao Duojiao (Lazy Cutie). She is a 1-year-old girl.}) \\ \midrule
Leisure & EN: What is the best order to watch films and tv shows in the MCU? \newline 
CN: \chinese{推荐一些比较容易翻唱的粤语歌给我}  (\textit{Recommend me some Cantonese songs that are easier to cover}) \\
\bottomrule
\end{tabular}
% }
\caption{Example English and Chinese Cases under each Intent. The notation "EN" represents that the case is in English. "CN" means the cases are reported in Chinese and we attach their English translation in italics. "......" represents the text behind is omitted. Note that due to space limitations, we present cases with shorter length in the above table. The average length of each question is 29.65 tokens in the URS dataset.}
\label{table:example}
\vspace{-2ex}
\end{table*}

% \section{Dataset}

% \section{User Reported Scenarios Dataset}
% \subsection{Dataset Construction}
\subsubsection{Construction}
\label{sec:construction}
In the initial phase of dataset construction, we check the diversity of our participants.
% , featuring 1863 authentic cases from 15 LLM services.
% The \textbf{U}ser \textbf{R}eported \textbf{S}cenario (URS) dataset is a compilation of scenarios actively reported by participants from the above user study. 
Figure~\ref{fig:IP} shows the distribution of automatically recorded IP for English and Chinese questionnaire respondents.
The feedback comes from 712 participants across 23 countries, showing the diversity in distributions. 
%This highlights the diversity of respondents. 
Detailed profiles, including demographic information~(age and occupation) and LLM usage experience, are provided in Appendix~\ref{appendix:participant}. 
% These highlight the extensive diversity of our respondents.

Subsequently, we examine the feedback on user intents.
There are no valid proposals under the "Others" option and manual fill-ins, suggesting a general comprehensive coverage of the proposed user intent taxonomy.

% Based on the high-quality feedback, we validate the user intent taxonomy and carry out multi-step data processing to construct the \textbf{U}ser \textbf{R}eported \textbf{S}cenario (URS) dataset.

Finally, we carry out meticulous data processing to construct the \textbf{U}ser \textbf{R}eported \textbf{S}cenario (URS) dataset as original feedback inevitably contains noise.
% The original feedback inevitably contains noise and thus we carry out meticulous data processing to obtain the high-quality URS dataset. 
%To form the high-quality URS dataset from original feedback, we entail meticulous data processing,
Steps include %encompassing 
format validation, elimination of intra-user duplication, and extraction of initial valid questions from each conversation. 
The dataset construction is augmented with third-party manual quality assessments to objectively filter out low-quality cases and confirm the exclusion of personal information. Detailed ethics considerations about this dataset are discussed in Section~\ref{section:ethical}.

% The construction process for URS dataset includes data cleaning~(format validation, removal of intra-user duplication), extraction of the initial valid query from each conversation and manually quality check for each question.

% Note that this dataset is composed of usage scenarios actively reported by users, there might exists certain selection bias, which is further discussed in Section~\ref{sec:limitation}.

\subsection{Dataset Information}
\label{sec:dataset_information}
% \noindent{\textbf{Statistics.}}
% This section presents the basic statistics of the URS dataset.
% emphasizing the variety of user intents and the distribution of LLM services which user conducted original conversations.
% \noindent \textbf{Statistic.}
Table~\ref{table:satistic_intents} reports the statistics of the URS dataset, with all data sourced from the above user study.
\begin{figure*}[t]
    \centering
    \includegraphics[width=\linewidth]{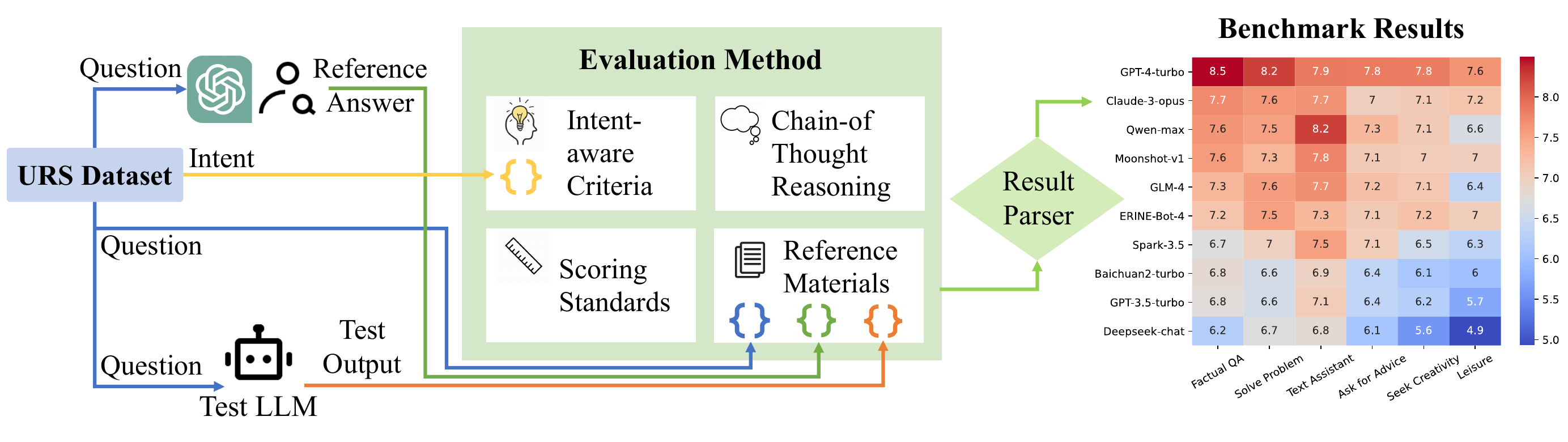}
    \caption{Evaluation Procedure. For each evaluation instance, the evaluator is provided with the user intent, five intent-aware criteria, chain-of-though reasoning steps, scoring standards for each two-point segment, addition with the question, an 8-score reference answer for this question, and the test LLM output for evaluation. Then, a parser will extract the final score from the evaluator's detailed rating content to form the benchmark.}
    \label{figure:framework}
\vspace{-2ex}
\end{figure*}
% \todo{remove intent analysis here}
% \noindent{\textbf{Intent Analysis.}}
% \todo{maybe remove}
% Based on the survey feedback, we categorized the 6 proposed intents into two groups: objective and subjective intents. This classification was guided by user requirements for personalization~\cite{bodonhelyi2024user}.
% The objective intents encompass user goals that involve seeking concrete answers or performing precise tasks where the results are fact-based, measurable, and usually have a concrete outcome (e.g., correct or wrong).
% In contrast, subjective intents involve user goals that may require personalized responses, guidance, or creative outputs where outcomes are influenced by personal situations or creative judgment.
% The intent analysis derived from the user study, which included 420 participants, revealed divergent expectations for personalization across the different intents. More than 55\% of participants expressed a preference for personalized feedback under the categories of Ask for Advice, Seek Creativity, and Leisure intents. Conversely, for the other intents, Factual QA, Solve Professional Problem, and Text Assistant, an average of less than 30\% of participants requested personalization. 
% encompassing both objective and subjective user intents.
%As detailed in Table~\ref{table:satistic_intents}, our proposed benchmark encompasses both objective~(Factual Question Answering, Solve Professional Problem, Text Assistant, Usage through API) and subjective~(Ask for Advice, Seek Creativity, Leisure) user intents.
In all intents, those that are relatively subjective, including Ask for Advice, Seek Creativity and Leisure, 
% Specifically, user intents of Ask for Advice, Seek Creativity and Leisure 
constitute about 33\% of the total use cases. 
As the traditional benchmarks primarily focus on evaluating LLMs' correctness of subjective questions~\cite{chang2023survey}, this objective portion of user needs is often overlooked.
%a portion of the users' needs that are often overlooked by traditional benchmarks, which primarily focus on evaluating LLMs' correctness of objective questions.
% with the latter often overlooked in ability-focused LLM benchmarks while accounting for 1/3 of usage.

% \begin{table}[t]
%     \centering
%     \resizebox*{0.5\textwidth}{!}{
%     \begin{tabular}{lr|lr}
% \textbf{Objective} &     & \textbf{Subjective} &      \\ \hline
% Factual QA         & 578 & Ask for Advice      & 275  \\
% Solve Problem      & 515 & Seek Creativity     & 191  \\
% Text Assistant     & 124 & Leisure             & 163  \\
% Usage through API  & 28  &                     &      \\ \hline
%                    &     & All                 & 1863
% \end{tabular}
%     }
%     \caption{Usage Distribution for each intent in real scenarios}
%     \label{table:satistic_intents}
% \end{table}

% \noindent{\textbf{LLMs Involved.}}
Besides the diversity in user intents,
the URS benchmark also includes interactions from 15 different LLM services. The distribution is detailed in Appendix~\ref{appendix:distribution_of_LLM}.
% This offers a more diverse range compared to similar user-centric benchmarks 
% that depend solely on logs from a single LLM service.
This breadth reduces potential biases stemming from reliance on a single LLM service. These biases can arise from users' limited access to certain LLMs and their preference for specific services in some scenarios.
Among this diversity, we observe a long-tail distribution in model usage. As we did not actively control or select use cases of certain LLMs during the dataset collection and construction processes, this may indicate the natural distribution in real-world usages.
% \todo{to check}
% \noindent \textbf{Time Factor.}
Time is also an important factor influencing LLM evaluations. Since the user study was conducted in February 2024, the collected cases in this dataset took places before this time.
% 1001	ChatGPT
% 54	ERNIE Bot
% 49	New Bing
% 33	Others
% 32	Bard
% 25	Gemini
% 16	ChatGLM
% 14	Character.AI
% 9	Claude
% 5	Baichuan
% 5	Qwen
% 5	Mistral
% 3	Llama
% 2	Spark
% 1	DeepSeek
% \todo{to be modified}

% \subsection{Cases}

% Table~\ref{table:example} illustrates the example cases. For each intent, we present one English question and on Chinese question translated into English for presentation. 

% \noindent \textbf{Example Cases.}
Table~\ref{table:example} showcases exemplary instances from the dataset. 
Real-world scenarios are often brief in articulation but complex in resolutions and span an expansive range of situations.
Notably, beyond linguistic differences, the cases in this dataset also encompass a variety of cultural backgrounds. These include traditional events, such as Major Snow and the Spring Festival in the lunar calendar as well as 
popular entertainment information that are globally or locally recognized, such as the Marvel Cinematic Universe and Cantonese songs. These multi-cultural contexts could enhance the benchmarking of LLMs, addressing more diverse needs of the global user base.

\section{URS Benchmark}
\subsection{Evaluation Framework}
\label{section:framework}
% \todo{overview picture}
We aim to design a evaluation framework that can autonomously, expeditiously, and precisely determine the performance of any LLMs.
% We first describe our evaluation criteria can be used %whose objective is to ensure autonomously, expeditiously, and precisely to determine the performance of an LLM. 
The automation is made possible by using a strong model (e.g., GPT-4) acting as the evaluator, whose effectiveness is validated in previous work for open-domain evaluations~\cite{chang2023survey,sottana2023evaluation,liu2023gpteval} and further examined in Section~\ref{section:sec:alignemt}.
% ~\cite{chang2023survey,lin2023llm,liu2023alignbench}
% with the Multi-dimensional Rule-Calibrated Judge
Aiming for high speed and precision, we implement a direct pair-wise scoring approach, providing a fixed reference answer for each question, in contrast to point-wise rating and Elo rating methods, as has been demonstrated to be effective in the previous study~\cite{alpaca_eval,liu2023alignbench}.
%The design objective is to ensure that when evaluating a new LLM, our benchmark can autonomously, expeditiously, and precisely assess its performance.
%To attain automation, we select a strong model to function as the base evaluator.
%For speed and precision, we implement a direct pair-wise scoring approach with reference answers provided, in contrast to point-wise rating and Elo rating methods. \todo{more reference!}
%In essence, we establish the Multi-dimensional Rule-Calibrated Judge~\cite{liu2023alignbench} with GPT-4.
As shown in Figure~\ref{figure:framework}, we categorize the evaluation instruction into 4 parts, including intent-aware criteria, chain-of-thought reasoning steps, scoring standards, and reference materials, which are described below % as follows
(The detailed instructions are provided in Appendix~\ref{appendix:instruction}):
%the evaluation instruction contains 5 parts: intent-aware criteria, chain-of-thought reasoning steps, scoring standards, reference materials, and required output format, as shown in Figure~\ref{figure:framework}.

\noindent \textbf{Intent-aware criteria.} 
During evaluations, we inform the evaluator about the user intent for each question and provide five specific evaluation criteria tailored to that intent. 
This approach is designed to help evaluators accurately capture user needs across various intent types and provide precise scorings.
% The evaluation instruction starts with an elucidation of user intent. Then, five corresponding criteria according to the specific intent and their explanations are included. 
Detailed illustrations of these criteria and the correspondence relationship between user intents and evaluation criteria are provided in Appendix~\ref{appendix:criteria}.
\noindent \textbf{Chain-of-thought reasoning steps.} 
To fully leverage the reasoning capabilities of evaluators, the instruction provides four steps before assigning final ratings.
% They are contrasting the reference answer with the test response under review, scoring each criterion on a 1-10 scale, re-thinking whether the response satisfied user needs, and amalgamating these assessments to formulate the final score. 
These steps include contrasting the reference answer with the test response, scoring each criterion on a 1-10 scale, reassessing whether the response meets user needs, and integrating these assessments to determine the final score.

\noindent \textbf{Scoring standards.}
To ensure accurate and differentiated scoring, we provide standards for each scoring segment.
% The instruction illustrates the score standards for each scoring bracket within the range from 1-10. 
Each 2-point increment constitutes a distinct segment. The answer furnished by us is an 8-score reference.
Besides, a note is provided to indicate that longer responses are not necessarily better to limit the potential length bias.
%This segment clarifies the benchmarks for each scoring bracket within the 1-10 range, with each two-point increment constituting a distinct category. The reference answer furnished by us is rated at 8.

\noindent \textbf{Reference materials.}
The instruction provided for the evaluator includes the question given by real users, their intent for asking, a fixed reference answer for this question~(provided by a strong LLM, which is GPT-4 in this paper and checked manually), and the response generated by the test LLM for evaluation.
%In this part, we give the query question, reference answer and the response awaiting evaluation.

% \noindent \textbf{Required output format.}
% The instruction requires the evaluator to produce the output encapsulated in JSON format, encompassing scores for the five intent-aware criteria and the overall score.
%The output should be encapsulated in JSON format, encompassing scores for the five intent-aware criteria and the overall score.

% \todo{evaluation method comparison}

% \begin{table*}[t]
%     \centering
%     \resizebox*{\textwidth}{!}{
%     \begin{tabular}{ll}
%     \toprule
%         Intents & Intent-aware Criteria \\ \midrule
%         Factual QA & Factuality, User Satisfaction, Clarity, Completeness, Logical Coherence \\
%         Solve Professional Problem & Factuality, User Satisfaction, Clarity, Logical Coherence, Completeness\\
%         Text Assistant & Clarity, User Satisfaction, Logical Coherence, Fatuality, Creativity\\
%         Usage through API & Factuality, User Satisfaction, Clarity, Logical Coherence, Completeness\\ \midrule
%         Ask for Advice & User Satisfaction, Factuality, Fairness and Responsibility, Creativity, Richness\\
%         Seek Creativity & User Satisfaction, Logical Coherence, Creativity, Richness, Factuality\\
%         Leisure & User Satisfaction, Engagement, Appropriateness, Creativity, Factuality\\ \bottomrule
%     \end{tabular} }
%     \caption{Evaluation Criteria under Different User Intents.}
%     \label{table:evaluation_criteria}
% \vspace{-2ex}
% \end{table*}

\begin{table*}[t]
\centering
\resizebox*{\textwidth}{!}{
\begin{tabular}{lrrrrrrr}
\toprule
% \multirow{2}{*}{\diagbox{Models}{Intents}} & Solve Problem & \makecell{Information\\Retrieval} & through API  & \makecell{Ask for\\Advice} & \makecell{Seek\\Creativity} & Leisure & All  \\ \cline{2-8} 
%                                              & 272           & 147                   & 26   & 58             & 57              & 60      & 620  \\ \hline
 & \makecell{Solve\\Problem} & \makecell{Factual\\QA} & \makecell{Text\\Assistant}  & \makecell{Ask for\\Advice} & \makecell{Seek\\Creativity} & Leisure & All  \\ \midrule
Cases  & 404  & 259 & 82   & 116 & 86  & 83  & 1024    \\ \midrule
GPT-4o & *8.24 & *8.49 & \textbf{7.86} & *7.85 & *7.83&  *7.65 & *8.15\\
% GPT-4-0125-preview & *8.30  & *8.68   & \textbf{7.92}  & *7.69 & *7.43  & *7.64        & *8.17                      \\ 
% GPT-4-turbo & 9.74 & 9.63 & 8.34 & 8.14 & 7.63 & 7.81 & 8.99 \\
Claude-3-opus-20240229 & \textbf{7.62}  & \textbf{7.73}  & 7.66   & 7.01   & 7.08    & \textbf{7.16}        & \textbf{7.50}  \\ 
Qwen-max & 7.53  & \underline{7.64}   & *8.24   & \textbf{7.28}   & 7.10    & 6.63        & \underline{7.48}                   \\ 
Moonshot-v1-8k         & 7.33  & 7.58   & \underline{7.77}   & 7.09   & 7.05    & 7.01        & 7.35                      \\ 
GLM-4 & \underline{7.55}  & 7.32   & 7.68 & \underline{7.20} & \underline{7.12}  & 6.37        & 7.33                      \\
ERNIE-Bot-4            & 7.54& 7.16  & 7.30 & 7.09 & \textbf{7.19}  & \underline{7.02}   & 7.31                      \\ 
Spark-3.5& 6.97& 6.71  & 7.53 & 7.05 & 6.54  & 6.33        & 6.87                      \\ 
Baichuan2-Turbo        & 6.59& 6.82  & 6.95 & 6.35 & 6.11  & 6.02        & 6.56                      \\ 
GPT-3.5-turbo          & 6.56& 6.76  & 7.05 & 6.35 & 6.24  & 5.75        & 6.53                      \\
Deepseek-chat          & 6.72& 6.22  & 6.81 & 6.09 & 5.56  & 4.93        & 6.29                      \\
   \bottomrule                 
\end{tabular}}
\caption{URS Benchmark Results. For each intent and the overall scenarios, we mark the three best-performing LLM services, with the first marked '*', the second bolded, and the third underlined.}
\label{table:result}
\vspace{-2ex}
\end{table*}

\subsection{Experimental Settings}
\label{section:setup}
% \noindent \textbf{LLM Services.}
Among the top-used LLM services reported in the previous user study, we test all the LLMs with available APIs, including 
Baichuan2-Turbo~\cite{Baichuan}, Claude-3-opus~\cite{Claude}, Deepseek-chat~\cite{bi2024deepseek}, ERNIE-Bot-4~\cite{ERNIE}, GLM-4~\cite{du2021glm}, GPT-3.5-turbo~\cite{ChatGPT}, GPT-4o~\cite{ChatGPT}, Moonshot-v1-8k~\cite{Moonshot}, Qwen-max~\cite{Qwen}, Spark-3.5~\cite{Spark}. We do not adjust the default temperature setting of LLMs based on query intent. This helps us to model the efficacy of LLM services in different real-world scenarios without user intent information.
We benchmark the above LLMs on 1024 human-examined cases in the URS dataset.

In the following sections, we conduct extensive experiments to answer three research questions:

\textbf{(RQ1)} What is the performance of different LLMs in the user-centric evaluation perspective? 

\textbf{(RQ2)} Are the evaluation results stable across different LLMs as evaluators? 

\textbf{(RQ3)} Do the benchmark results align with real user perceptions? 
%They are listed below in alphabetic order:

% - Baichuan2-Turbo~\cite{Baichuan},

% - Claude-3-opus~\cite{Claude}, 

% - Deepseek-chat~\cite{Deepseek},

% - ERNIE-Bot-4~\cite{ERNIE},

% - GLM-4~\cite{GLM},

% - GPT-3.5-turbo~\cite{ChatGPT},

% - GPT-4-0125-preview~\cite{ChatGPT},

% - Moonshot-v1-8k~\cite{Moonshot},

% - Qwen-max~\cite{Qwen},

% - Spark-3.5~\cite{Spark}.

% They are all close-sourced LLM services.

\subsection{Overall Results (RQ1)}
\label{sec:overall_result}

The overall benchmark results are shown in Table~\ref{table:result}. We have the following observations:

(1) GPT-4 yields an average score of 8.15 when its previously generated responses are used as 8-score references in the scoring instruction. This difference indicates an acceptable level of variance from the established standard. The benchmark stability across different evaluators is further discussed in the next section.
% When the responses of GPT-4 are used as references in the scoring instruction, i.e., equating to a score of 8, the subsequent assessments of the re-generated GPT-4 results yield an average score of 8.15, indicating an acceptable level of variance from the established standard.
%In our scoring instruction, GPT-4's responses are used as references, equating to a score of 8. Subsequent assessments of the re-generated GPT-4 results yield an average score of 8.16, indicating an acceptable level of variance from the established standard. \todo{further evaluation}

(2) There is a noticeable stratification in scores across different models. Apart from GPT-4, the leading group comprises Claude-3 and Qwen-max, followed by a secondary tier including Moonshot-v1, GLM-4, and ERNIE-Bot-4. The performances of the third tier are comparable to that of GPT-3.5. 
% This suggests a proper differentiation ability of our URS benchmark.

(3) In examining efficacy across diverse user intents, GPT-4 generally demonstrates superior performance, except for the Text Assistant category, where Qwen-max prevails.

Furthermore, a comparative analysis of LLMs across different user intents reveals a pronounced proficiency in objective contexts, including Solve Problem and Factual QA. This is probably because subjective scenarios require more diverse competencies such as personalization, creativity and humor, where current LLMs are relatively weak.

Separated results of English and Chinese scenarios are presented in Appendix~\ref{appendix:results_EN_CN}. Scoring examples are provided in Appendix~\ref{appendix:scoring_example}.

\begin{figure}[t]
    \centering
    \includegraphics[width=\linewidth]{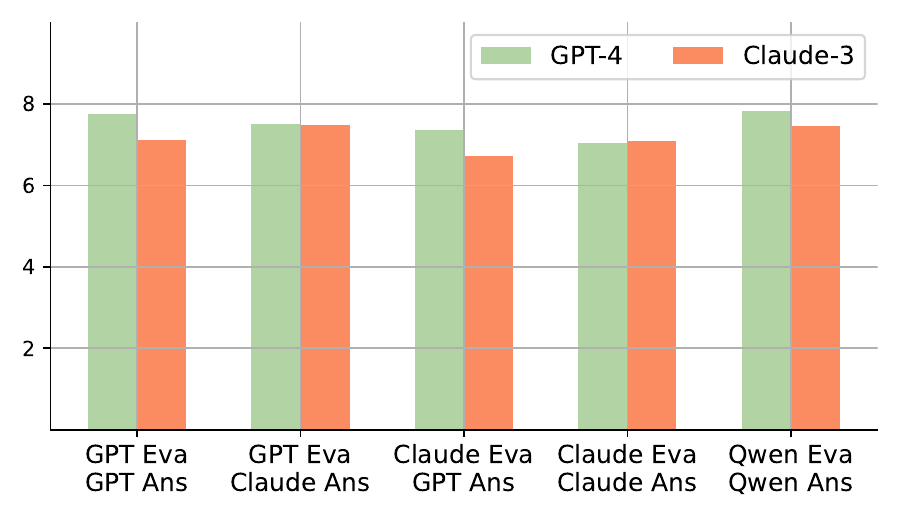}
    \caption{Comparison between GPT-4 and Claude-3. "A Eva B Ans" indicates the evaluation setting, where 'A' denotes the evaluator LLM and 'B' represents the source LLM used to generate 8-point reference answers. These results show that GPT-4 slightly outperforms Claude-3.}
    \label{fig:gpt-claude}
\vspace{-2ex}
\end{figure}

\subsection{Cross Validation between GPT-4 and Claude-3 (RQ2)}
\label{sec:cross_validation}
In the overall benchmark result, we adopt GPT-4's direct output to the question as a reference for score 8, and use GPT-4 as the evaluator base for final scoring, as detailed in Setion~\ref{section:framework}. Although this approach is widely adopted, it may cause favor to responses similar to GPT-4. To counteract this bias, we re-evaluated the performance of the top-2 LLMs, GPT-4o, and Claude-3-opus, employing cross-validation techniques.
We run 5-fold evaluation settings on randomly selected 200 cases, using GPT and Claude as the reference generator and base evaluator respectively, and the third performed LLM, Qwen, as both reference generation and evaluation for a third-party evaluation.
As illustrated in Figure~\ref{fig:gpt-claude}, 
when GPT-generated responses are used as references, GPT-4 consistently achieves higher scores no matter the evaluator models. When Claude-generated responses are adopted in evaluations, the results are comparable.
% For the first 4 settings, GPT-4 scores higher with GPT generated answers and scores comparable with Claude as answers. 
In third-party assessments, GPT-4 also maintained a superior ranking over Claude-3.
This order is consistent with the overall benchmark results, indicating that the potential bias towards GPT does not affect the benchmark ranking.
% and the latest released gpt-4-turbo with cross-validation methods.

\subsection{Alignment with Human Experience~(RQ3)}
\label{section:sec:alignemt}

% We compare the benchmark scores with participants' reported general satisfaction in LLM interactions across different intents, collected in Section~\ref{section:user_study}.

To further evaluate the efficacy of LLM-as-Judge design in this benchmark, we analyzed benchmark accuracy through the lens of real-world human experience, examining both intent-wise user satisfaction and pairwise LLM-wise comparisons.

\begin{figure}[t]
    \centering
    \includegraphics[width=\linewidth]{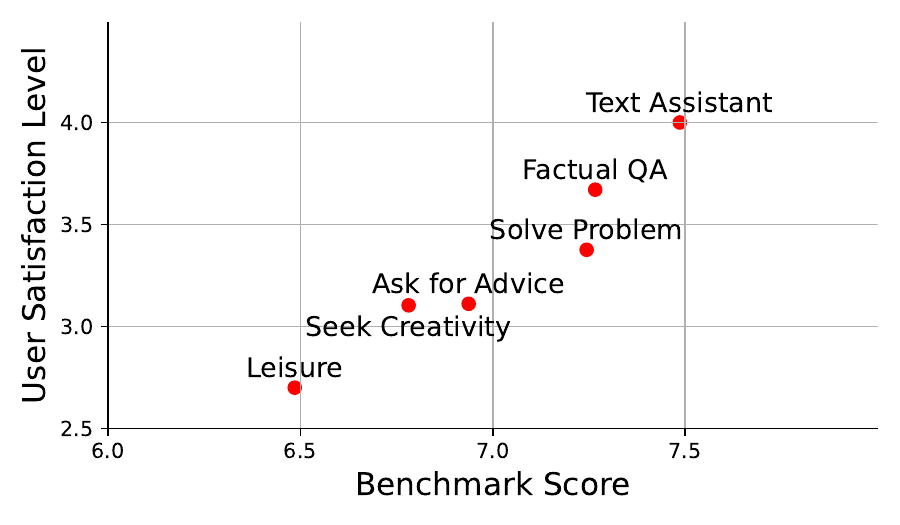}
    \caption{Benchmark Score and User Reported Satisfaction Correlate Well across Intents. "Benchmark Score" is averaged under different intents. "User Reported Satisfaction" is the average satisfaction level reported in the user study. Intents are ranked by user satisfaction.}
    \label{figure:alignment_intent}
\vspace{-2ex}
\end{figure}

For intent-wise evaluation, user satisfaction levels were collected in the user study detailed in Section~\ref{section:user_study}, involving 420 global participants who rated their satisfaction on a five-point scale for each intent while using LLM services. A high Pearson correlation coefficient of 0.95 between the average benchmark score and user-reported satisfaction across different intents, as shown in Figure~\ref{figure:alignment_intent}, confirms the alignment.
Notably, subjective scenarios such as Ask for Advice, Seek Creativity and Leisure received lower ratings than more subjective ones in both benchmark results and user feedback. This indicates the current LLM services still need improvements in satisfying real-world user needs.
% There is a high Pearson correlation coefficient of 0.95 between the average benchmark score and user-reported satisfaction across different intents, as depicted in Figure~\ref{figure:alignment_intent}. Note that the two groups are not directly comparable due to scale differences, while the relevant ordering of intents is consistent across both measures.
% The results reveal that objective scenarios, such as Ask for Advice, Seek Creativity and Leisure, consistently receive higher ratings than subjective ones in both the benchmark results and user feedback. 
% Additionally, user-reported satisfaction displays more pronounced variations between the highly and poorly performing intents compared to the benchmark scores. 
% This suggests that users are more sensitive to differences in service performance, underscoring the need to further incorporate this characteristic into evaluation designs to more precisely capture user experience.

% \begin{figure}[t]
%   \centering
%   \begin{subfigure}[b]{0.45\linewidth}
%     \includegraphics[width=\textwidth]{figure/Alignment-Intent.pdf}
%     \caption{First Image}
%     \label{fig:sub1}
%   \end{subfigure}
%   \hfill % 添加空格
%   \begin{subfigure}[b]{0.45\linewidth}
%     \includegraphics[width=\textwidth]{figure/Alignment-LLM.pdf}
%     \caption{Second Image}
%     \label{fig:sub2}
%   \end{subfigure}
%   \caption{Images of example}
%   \label{fig:test}
% \end{figure}

\begin{figure}[t]
    \centering
    \includegraphics[width=\linewidth]{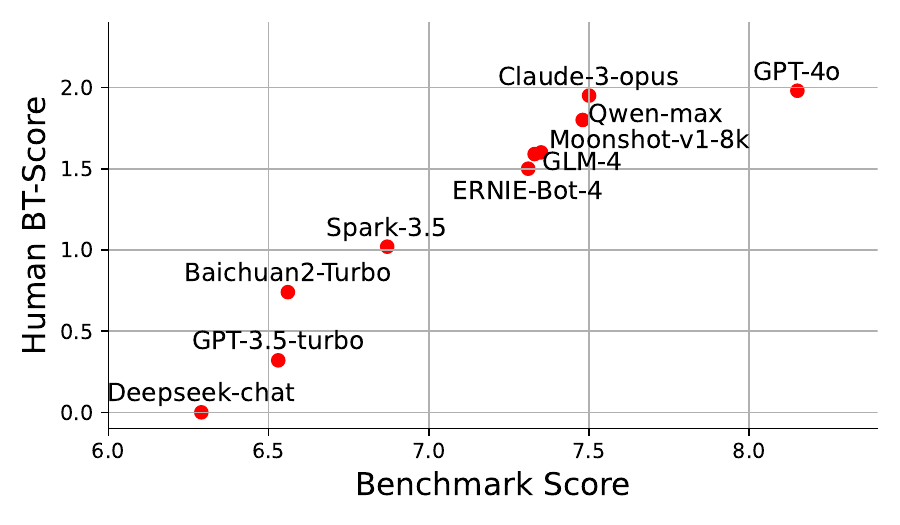}
    \caption{LLMs' Benchmark Ranking Aligns with the Order from Human Pairwise Annotation.}
    \label{figure:alignment_LLM}
\vspace{-2ex}
\end{figure}
% \begin{table}[]
% \begin{tabular}{llll}
%                        & \makecell{Benchmark\\Evaluation}         & \multicolumn{2}{l}{\makecell{Human\\Preference}}                     \\
%                        & Overall Score                & Win Rate                     & BT-Score                     \\
% gpt-4o                 & \cellcolor[HTML]{F8696B}8.15 & \cellcolor[HTML]{F8696B}0.59 & \cellcolor[HTML]{F8696B}1.98 \\
% claude-3 & \cellcolor[HTML]{FECF7F}7.50 & \cellcolor[HTML]{F9706D}0.59 & \cellcolor[HTML]{F9726D}1.95 \\
% qwen-max               & \cellcolor[HTML]{FED380}7.48 & \cellcolor[HTML]{FA7F70}0.58 & \cellcolor[HTML]{FB9F76}1.80 \\
% moonshot-v1-8k         & \cellcolor[HTML]{FFE784}7.35 & \cellcolor[HTML]{FB9875}0.56 & \cellcolor[HTML]{FFDA81}1.60 \\
% glm-4                  & \cellcolor[HTML]{FFEA84}7.33 & \cellcolor[HTML]{FFE182}0.52 & \cellcolor[HTML]{FFDD82}1.59 \\
% ERNIE-Bot-4            & \cellcolor[HTML]{FDEA83}7.31 & \cellcolor[HTML]{FCEA83}0.51 & \cellcolor[HTML]{FAE983}1.50 \\
% spark-3.5              & \cellcolor[HTML]{BAD780}6.87 & \cellcolor[HTML]{CDDC81}0.40 & \cellcolor[HTML]{C9DB80}1.02 \\
% Baichuan2-Turbo        & \cellcolor[HTML]{8CC97D}6.56 & \cellcolor[HTML]{A0CF7E}0.30 & \cellcolor[HTML]{AED37F}0.74 \\
% gpt-3.5-turbo          & \cellcolor[HTML]{87C87D}6.53 & \cellcolor[HTML]{83C77C}0.24 & \cellcolor[HTML]{83C77C}0.32 \\
% deepseek-chat          & \cellcolor[HTML]{63BE7B}6.29 & \cellcolor[HTML]{63BE7B}0.17 & \cellcolor[HTML]{63BE7B}0.00
% \end{tabular}
% \end{table}

In addition to the above alignment analysis based on large-scale user studies, we further conducted pair-wise annotations according to methods in Chatbot Arena~\cite{chiang2024chatbot} to check the consistency between the designed automatic evaluation and human preferences. Due to space limitations, the detailed approach is provided in Appendix~\ref{appendix:alignment}. Based on 400 paired comparisons assessed by 5 human annotators, we computed Bradley-Terry coefficients~\cite{bradley1952rank} for each LLM. As shown in Figure~\ref{figure:alignment_LLM}, the ranking of LLMs according to these coefficients aligns with the benchmark rankings with Pearson r=0.94, further affirming the effectiveness of evaluation designs.

\section{Conclusion}
By focusing on real-world cases, user intent categorization, and intent-aware evaluation, this work endeavors to advance the user-centric benchmark of LLMs. 
We hope the insights gained from this research will not only help improve the performance of LLMs but also foster a deeper understanding of how these services can be effectively integrated into people's lives to enhance productivity, creativity, and overall welfare.

\section*{Limitations}
\label{sec:limitation}

While evaluating LLMs from the user-centric perspective offers considerable advantages, there are certain limitations in our URS benchmark.

Evaluation Method: Our approach aligns with prevailing practices in open-ended question evaluation, where GPT-4 is employed as the base evaluator. This could potentially lead to a bias toward responses resembling GPT’s style, thereby influencing the neutrality of the evaluation. We analyzed through cross-validation that this approach does not affect the top 2 LLM orders. Besides, analysis shows that benchmark score aligns with real-world user experience and human preferences in pairwise annotations.
These further tested the validity of our method.
In addition, the user study and human pairwise annotations could be further scale up to enhance persuasiveness.

Besides, we notice it is very important to evaluate LLM services' performances in multi-turn conversational settings, as there are a few cases where the LLMs do not answer the user question and ask to provide more detailed information about user intents. These may be judged unsuccessful in a single-turn assessment, but the service might be capable of meeting this user need.

Note that this benchmark is not holistic, as we mainly measure LLMs' efficacy in satisfying real-world user needs.
User satisfaction is indeed crucial, while it represents one aspect of LLM performance. Other critical facets include effectiveness, efficiency, ease of use, and error tolerance. In this study, our primary focus is on user-centric evaluation of LLMs, which we have validated through strong alignments with human preferences to address its effectiveness.

Data Distribution: The data for our study was mainly sourced from a global researcher recruitment platform, which does not ensure a random selection of all participants using LLM services. Consequently, the demographic distribution of our dataset may not accurately reflect that of typical LLM users or the global population. 
While as an independent third party unaffiliated with specific LLM services, and given our non-selective approach to data collection, our methodology incorporates greater diversity and offers a closer alignment with real-world usage scenarios compared to datasets that are
human-designed, model synthetic, or selected by non-third parties or single-sourced.
% As we are a third-party without relationships to certain LLM services and do not manually selected the collected logs, this approach contains more diversity and aligns more with real-world scenarios compared to human-written datasets or non-third-party selection of logs from certain LLM services.

% Despite the multiple advantages of evaluating LLMs from a user-centric perspective,
% it demonstrates two limitations for benchmarking LLM.

% Evaluation Method.
% For the evaluation method, we follow most work for open-end quality evaluation to choose GPT-4 as the evaluator. This might cause an incline to answer similar to GPT's output.
% In the evaluation, we tested the \todo{show 2 fold experiment results}.

% Data Distribution
% The questionnaire was collected through a global recruitment platform rather than a strictly random selection of subjects.
%  The distribution of populations involved in the dataset deviates from the distribution of populations using LLM, or the distribution of the world's population

\section*{Ethical Considerations}
\label{section:ethical}
In the user study, participants were informed that their reported conversations would be publicly released for research purposes. The notification is detailed in Appendix~\ref{sec:appendix}. At the same time, they voluntarily chose whether to provide anonymized basic information, such as age group and profession.

Additionally, the survey design included an anchor question requiring participants to select specific options; responses that failed to adhere to these instructions were discarded as malicious feedback.

The questionnaire was provided in English and Chinese~(Mandarin), the top 2 widely spoken languages\footnote{\url{https://en.wikipedia.org/wiki/List_of_languages_by_total_number_of_speakers}}, and distributed via the global recruitment platform\footnote{\url{https://www.prolific.com}}. In this process, we did not filter any attributes as this could actively introduce bias. Following data collection, a third party conducted manual reviews to ensure the data's quality, harmlessness, and the absence of any personal information.

This dataset could be used for research purposes to understand user intents and real-world needs, evaluate conversational systems' responses, and train for better alignment with human preference.
Note that this URS dataset uses an Apache License.

%% file: section/appendix.tex
\clearpage
\section*{Appendix}
\section{User Study}
\label{sec:appendix}
%Here are the questions in the user survey.
At the start of the user survey, we have this notification:

\textit{Your dialog reported in this questionnaire may be publicly released for research purposes, so please avoid including personal information, and we will also conduct a secondary check, thank you!}

Subsequently, participants are asked to provide five to ten distinct real use cases with LLM services. Each use case should comprehensively cover the following aspects:

1. LLM Service Utilized: \\
\textit{Which of the following large language models did you ask the question to}

2. Conversation Content:\\
% \noindent Description: Detailing the conversations through copying of texts or sharing links. \\
\textit{Your question and the system's output \\
It is recommended to directly share your conversation link!\\
If you fill it out manually, please respectfully follow the format below.\\
Question: xxx\\
Answer: xxx}

3. Usage Intent:\\
% \noindent Description: Clarifying the user intent behind starting this conversation. \\
\textit{Your intent behind this question is}

% 4. Task Difficulty:\\
% % \noindent Description: Assessing the complexity level of the tasks handled by the LLM. \\
% \textit{How difficult do you think the problem is}

% 5. Satisfaction Level:\\
% % \noindent Description:  Rating their satisfaction with the LLM's response.\\
% \textit{Your satisfaction with the system output}

Participants are also asked to voluntarily report the following demographic information anonymously:

1. Age Group: To understand the age distribution of LLM users.

2. Profession: To gauge the professional background and its influence on LLM usage.

3. IP Location: To capture the geographical distribution of participants.

General usage Experience with LLMs:

1 LLM used: \\
\textit{Large Language Models you have used }

2 Usage Frequency: \\
\textit{How often do you use LLMs}

3 Satisfaction with LLMs:\\
\textit{Your general satisfaction with LLM services across each intent}

\section{URS Dataset Information}
\subsection{Participants}
\label{appendix:participant}
% Note that this URS dataset uses an Apache License.

\begin{figure}[t]
    \centering
    \includegraphics[width=\linewidth]{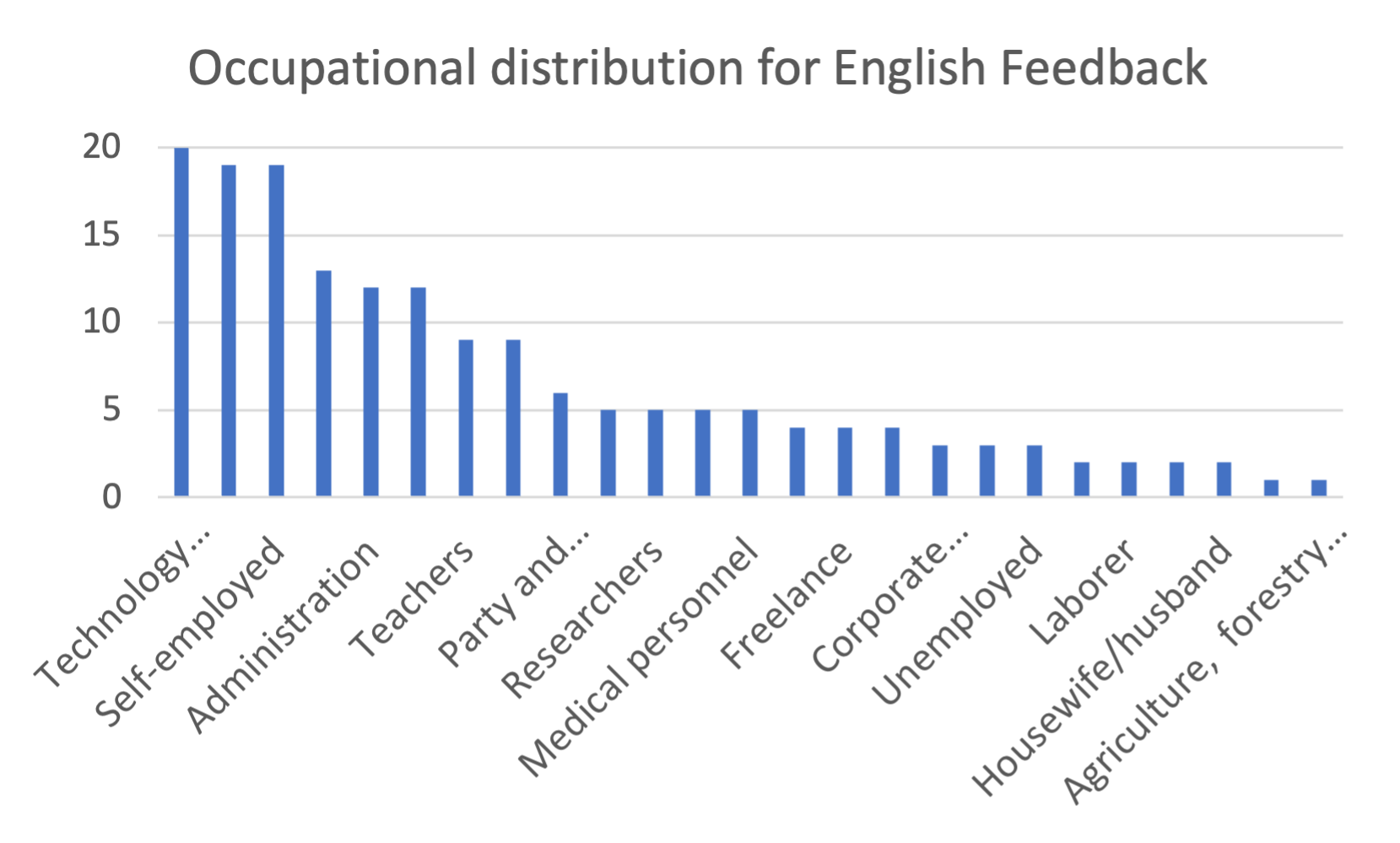}
    \includegraphics[width=\linewidth]{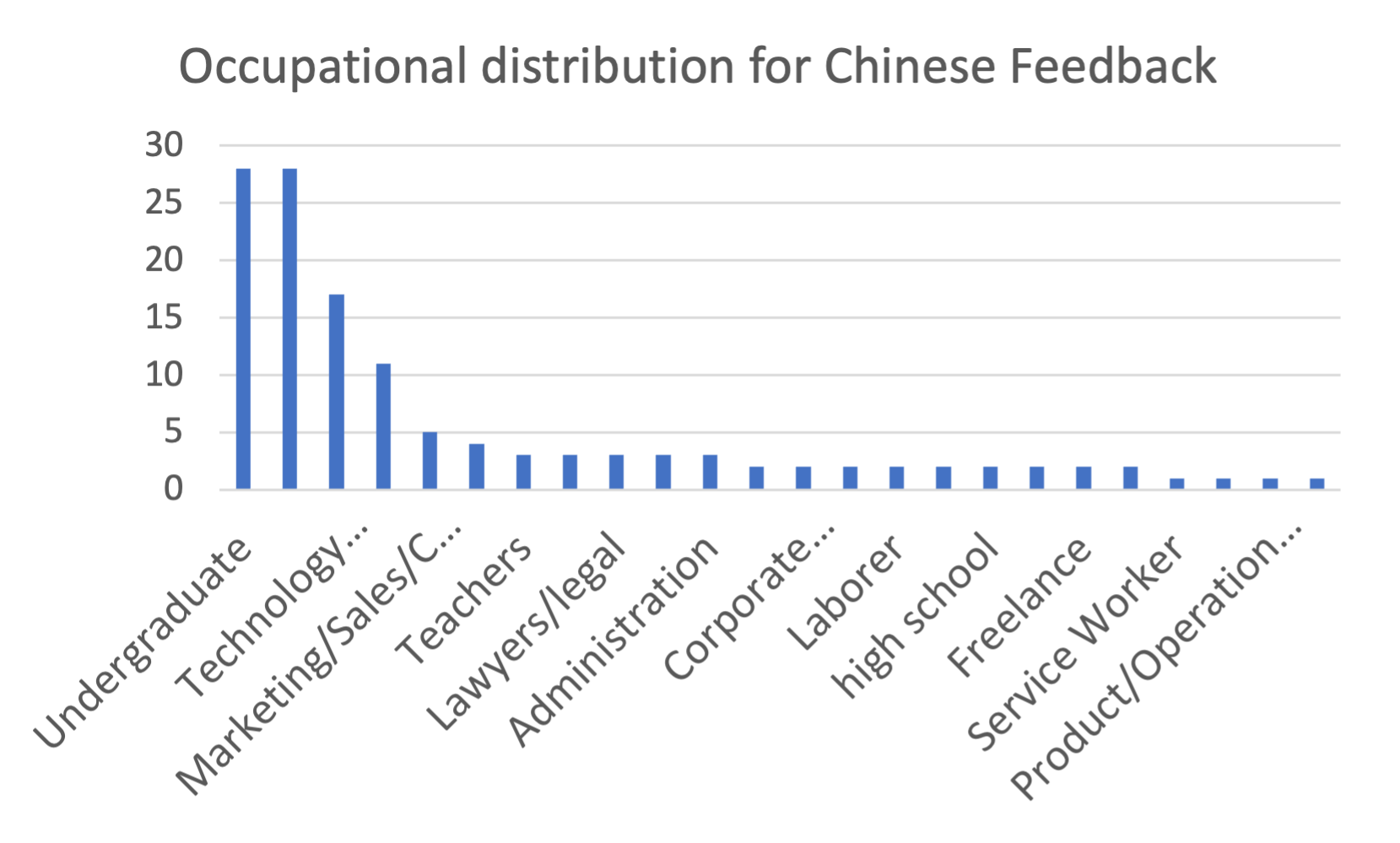}
    \caption{Occupation distribution of survey participants.}
    \label{fig:occupation}
\end{figure}

\begin{figure}[t]
    \centering
    \includegraphics[width=\linewidth]{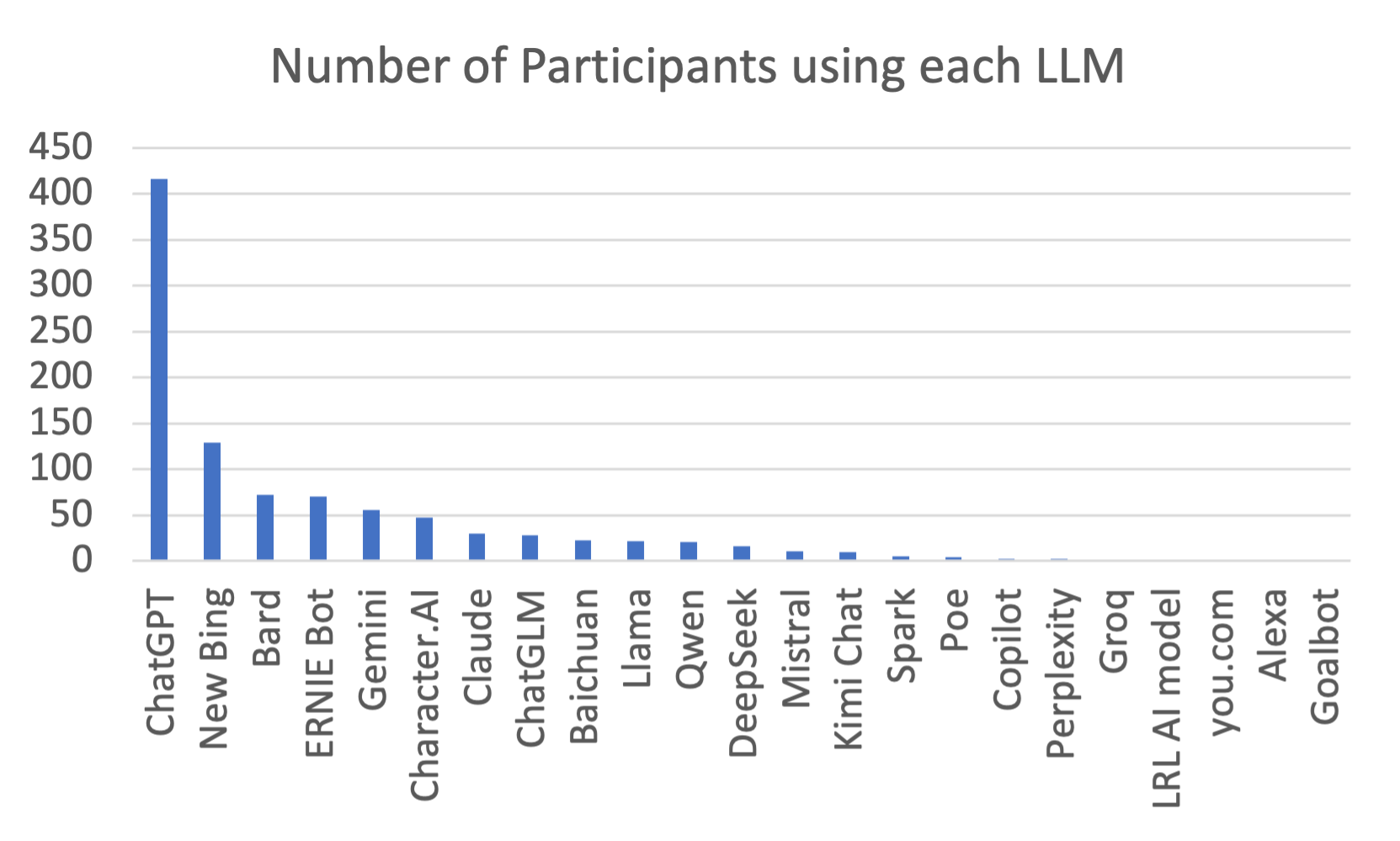}
    \caption{Number of Participants reported having used each LLM.}
    \label{fig:user-llm}
\end{figure}

Figure~\ref{fig:occupation} and Figure~\ref{fig:user-llm} report the demographic information and the LLM services usage pattern of the user study participants, which demonstrates the diversity of the benchmark contributors.

%As shown in Figure~\ref{fig:occupation} and Figure~\ref{fig:user-llm}, we demonstrate the demographic information and the usage pattern with LLMs to illustrate the diversity of our participants.

\subsection{Distribution of LLM usage}
\label{appendix:distribution_of_LLM}
Table~\ref{table:statistic_distribution} presents the distribution of LLM usage from the collected user study. ChatGPT is the most popular one in both language backgrounds, while the others show some differences.
\begin{table}[t]
    \centering
    \begin{tabular}{lrlr}
        \toprule
        \textbf{English} & & \textbf{Chinese} & \\ \midrule
        Total & 993 & Total & 994 \\ \midrule
        ChatGPT & 809 & ChatGPT & 818 \\
        Bard/Gemini & 69 & ERNIE Bot  & 61 \\
        New Bing & 41 & New Bing & 41 \\
        Character.AI & 14 & Baichuan & 7  \\
        ChatGLM & 7 & Character.AI, etc. & 6  \\
        \bottomrule
    \end{tabular}
    \caption{Distribution on Top 5 Source LLMs. In the URS dataset, most conversations are reported with their source LLMs. Notely, for Chinese feedback, Character.AI, Claude, Mistral, and Qwen are tied for 5th place.}
    \label{table:statistic_distribution}
\end{table}

\section{Evaluation}
\subsection{Intent-aware Criteria}
\label{appendix:criteria}

\begin{table*}[t]
    \centering
    \resizebox*{\textwidth}{!}{
    \begin{tabular}{ll}
    \toprule
        Intents & Intent-aware Criteria \\ \midrule
        Factual QA & Factuality, User Satisfaction, Clarity, Completeness, Logical Coherence \\
        Solve Professional Problem & Factuality, User Satisfaction, Clarity, Logical Coherence, Completeness\\
        Text Assistant & Clarity, User Satisfaction, Logical Coherence, Fatuality, Creativity\\ \midrule
        % Usage through API & Factuality, User Satisfaction, Clarity, Logical Coherence, Completeness\\ \midrule
        Ask for Advice & User Satisfaction, Factuality, Fairness and Responsibility, Creativity, Richness\\
        Seek Creativity & User Satisfaction, Logical Coherence, Creativity, Richness, Factuality\\
        Leisure & User Satisfaction, Engagement, Appropriateness, Creativity, Factuality\\ \bottomrule
    \end{tabular} }
    \caption{Evaluation Criteria under Different User Intents.}
    \label{table:evaluation_criteria}
\vspace{-2ex}
\end{table*}

\begin{table}[t]
    \centering
    \begin{tabular}{lrr}
    \toprule
    Model & Max L & Accuracy \\ \midrule
    Claude-3-opus     & 19& 100\% \\
    GPT-4     & 12& 100\%\\
    ERNIE-Bot-4 & 9 & 94\%\\
    GPT-3.5-turbo & 8 & 86\% \\
    Qwen-max & 4 & 57\% \\
    spark-3.5 & 4 & 50\% \\
    \bottomrule
    \end{tabular}
    \caption{Addition Ability. "Max L" represents the longest sequence of consecutive additions attainable at a minimum accuracy rate of 80\%. "Accuracy" denotes the success ratio in 90 trials covering the length of 2 to 9 with 10 test cases of each length. Outcomes are presented for LLMs with an accuracy of 50\% and above.}
    \label{table:addition}
\end{table}

The definition of intent-aware criteria is shown in Table~\ref{fig:criteria_description}. We evaluate each defined user intent with five selected criteria as described in Table~\ref{table:evaluation_criteria}.

\begin{table*}[t]
\begin{tabular}{{p{3.5cm}p{11.5cm}}}
\toprule
Criteria                               & Description   \\ \midrule
 Factuality \newline \chinese{事实正确性}                     & 
 Whether the information provided in the response is accurate, based on reliable facts and data. \newline \chinese{回答中提供的信息是否准确无误，是否基于可信的事实和数据。}                                                                                          \\
User Satisfaction \newline \chinese{满足用户需求}              &  Whether the response meets the user’s question and needs, and provides a comprehensive and appropriate answer to the question.  \newline  \chinese{回答是否满足了用户提出问题的目的和需求，是否对问题进行了全面而恰当的回应。}                                              \\
 Logical Coherence   \newline \chinese{逻辑连贯性}       & Whether the response maintains overall consistency and logical coherence between different sections, avoiding self-contradiction.  \newline \chinese{回答是否在整体上保持一致，是否在不同部分之间保持逻辑连贯性，避免了自相矛盾。}                                           \\
 Richness \newline \chinese{丰富度}     & Whether the response includes rich info, depth, context, diversity, detailed explanations, and examples to meet user needs and provide a comprehensive understanding. \newline  \chinese{回答包含丰富的信息、深度、上下文考虑、多样性、 详细解释和实例，以满足用户需求并提供全面理解。}  \\  \midrule
 Creativity   \newline \chinese{创造性}                     &  Whether the response is innovative or unique, providing novel insights or solutions.  \newline  \chinese{回答是否具有创新性或独特性，是否提供了新颖的见解或解决方法。}                                                                                            \\
Fairness and Responsibility \newline \chinese{公平与可负责程度} & Whether the advice or information provided in the response is feasible, carries a certain degree of responsibility and considers potential risks and consequences. \newline  \chinese{回答中提供的建议或信息是否可行， 是否负有一定的责任，是否考虑了潜在风险和后果。}         \\  \midrule
 Completeness \newline    \chinese{完备性}                & Whether the response provides sufficient information and details to meet the user’s needs, and whether it avoids omitting important aspects.  \newline \chinese{回答是否提供了足够的信息和细节，以满足用户的需求，是否遗漏了重要的方面。}                                \\
Clarity \newline          \chinese{清晰度}               & Whether the response is clear and understandable, and whether it uses concise language and structure so that the user can easily understand it. \newline          \chinese{回答是否表达清晰易懂，是否使用了简洁的语言和结构，以便用户可以轻松理解。}                               \\
Engagement \newline \chinese{趣味性}                &  Whether the answer is interesting and attractive, helps users relax, and provides high-quality emotional value or entertainment value, etc.\newline         \chinese{回答是否有趣、吸引人，帮助用户放松，提供了高质量的情绪价值或娱乐价值等}                                                                                                                                                                               \\
Appropriateness \newline \chinese{适宜性}             & Content is suitable for all users and avoids inappropriate or offensive content.  \newline   \chinese{内容适宜所有用户，避免不当或冒犯性内容} \\ \bottomrule
\end{tabular}
\caption{Definition of Intent-aware Criteria.}
\label{fig:criteria_description}
\end{table*}

\subsection{Instruction}
\label{appendix:instruction}
Here is the template for each evaluation case.
\begin{tcolorbox}[title = {Instruction Template},breakable]
\small
You are asked to assess the quality of an AI assistant's answer to a user's question as an impartial judge. Since the type of answer you are evaluating is [\{user\_intent\}], you need to evaluate the answer in the following 5 criteria:

1 \{criteria\_dict[user\_intent][0]\}

2 \{criteria\_dict[user\_intent][1]\}

3 \{criteria\_dict[user\_intent][2]\}

4 \{criteria\_dict[user\_intent][3]\}

5 \{criteria\_dict[user\_intent][4]\}\\

We will provide you with the user's question, an 8-score reference answer, and answers from the AI assistant that needs your assessment. When starting your evaluation, you need to follow the reasoning steps below:\newline
1. Compare the AI assistant's answer with the reference answer, point out any shortcomings in the AI assistant's answer, and explain further. \newline
2. Evaluate the AI assistant's answer in terms of the different criteria, giving each criterion a score from 1 to 10 after the evaluation of each. \newline
3. Finally, combine the evaluations from each criterion and give the AI assistant's answer a composite score of 1 to 10. \newline
4. Your scoring needs to be as rigorous as possible and adhere to the following scoring rules: in general, the higher the quality of the model's answers, the higher the score. \\ 

When the model answer has irrelevance to the question, or intrinsically factually incorrect, or generates harmful content, the total score should be 1 to 2;\newline
When the model answer has no serious errors and is largely harmless, but is of low quality and does not meet user requirements, the total score must be 3 to 4; \newline
When the model answer basically meets the user's needs but performs poorly on some criteria and is of medium quality, the total score can be 5 to 6; \newline
When the quality of the model response is similar to the reference answer and performs well in all criteria, the total score should be 7 to 8; \newline
A score of 9 to 10 can only be achieved if the model significantly exceeds the quality of the reference answer, adequately addresses the user's question and all the needs, and is close to a perfect score on all criteria. 
As an example, the reference answer would receive a score of 8. \\ 
Do not let the length of the answer affect your score. Longer answers are not necessarily better. Answers that are concise and meet the above requirements are good.\\

You need to evaluate and explain before you score. Your explanation of each criterion needs to be followed by the scoring. After that, at the end of your answer, return all of your scores in the following dictionary format, including the curly brackets, and make sure that your scores are integers: \newline
\{'Dimension 1': score, 'Dimension 2': score, ... , 'Final Score': score\}, e.g. \{'\{criteria\_dict[user\_intent][0]\}': 9, '\{criteria\_dict[user\_intent][1]\}': 6, ... , 'Final Score': 7\}. \\ 

User Question: {question} \newline
 [Start of Reference Answer] \newline \{reference\_ans\} \newline[End of reference answer] \newline
 [start of AI assistant's answer]\newline \{test\_model\_output\} \newline[end of AI assistant's answer

\end{tcolorbox}

The instructions for evaluation with four parts are shown in Table~\ref{tab:English_instruction}, ~\ref{tab:English_instruction_v2} and Table~\ref{tab:chinese-instruction} in English and Chinese, respectively. We use these designed instructions to guide the evaluator to output the evaluation results for benchmarking.

% You are asked to assess the quality of an AI assistant's answer to a user's question as an impartial judge. Since the type of answer you are evaluating is [\{user\_intent\}], you need to evaluate the answer in the following 5 criteria:

% We will provide you with the user's question, an 8-score reference answer, and answers from the AI assistant that needs your assessment. When starting your evaluation, you need to follow the reasoning steps below:\newline
% 1. Compare the AI assistant's answer with the reference answer, point out any shortcomings in the AI assistant's answer, and explain further. \newline
% 2. Evaluate the AI assistant's answer in terms of the different criteria, giving each criterion a score from 1 to 10 after the evaluation of each. \newline
% 3. Finally, combine the evaluations from each criterion and give the AI assistant's answer a composite score of 1 to 10. \newline
% 4. Your scoring needs to be as rigorous as possible and adhere to the following scoring rules: in general, the higher the quality of the model's answers, the higher the score.

\begin{table*}[t]
    \centering
    \begin{tabular}{{p{1.5cm}p{13.5cm}}}
    \toprule
     Intent-aware \newline Criteria    & You are asked to assess the quality of an AI assistant's answer to a user's question as an impartial judge. Since the type of answer you are evaluating is [Solve Professional Problem], you need to evaluate the answer in the following 5 criteria: \newline
1 Factuality \newline
Whether the information provided is accurate and based on reliable facts and data \newline
2 User Satisfaction \newline
Whether the response meets the user’s question and needs and provides a comprehensive and appropriate answer to the question \newline
3 Clarity \newline
Whether the response is clear and understandable, and whether it uses concise language and structure so that the user can easily understand it. \newline
4 Logical Coherence \newline
Whether the response maintains overall consistency and logical coherence between different sections, avoiding self-contradiction \newline
5 Completeness \newline
Whether the response provides sufficient information and details to meet the user’s needs, and whether it avoids omitting important aspects \newline
Note that a longer answer is not always better, the answer that is concise and meets the above requirements is the best. \\ \midrule
Chain of  \newline thought \newline reasoning steps     & We will provide you with the user's question, an 8-score reference answer, and answers from the AI assistant that needs your assessment. When starting your evaluation, you need to follow the reasoning steps below:\newline
1. Compare the AI assistant's answer with the reference answer, point out any shortcomings in the AI assistant's answer, and explain further. \newline
2. Evaluate the AI assistant's answer in terms of the different criteria, giving each criterion a score from 1 to 10 after the evaluation of each. \newline
3. Finally, combine the evaluations from each criterion and give the AI assistant's answer a composite score of 1 to 10. \newline
4. Your scoring needs to be as rigorous as possible and adhere to the following scoring rules: in general, the higher the quality of the model's answers, the higher the score. \newline
The two most important criteria are factual correctness and fulfillment of user needs, and the scores for these two dimensions dominate the final composite score. \\ \midrule
Scoring \newline Standards & When the model answer has irrelevance to the question, or intrinsically factually incorrect, or generates harmful content, the total score should be 1 to 2;\newline
When the model answer has no serious errors and is largely harmless, but is of low quality and does not meet user requirements, the total score must be 3 to 4; \newline
When the model answer basically meets the user's needs but performs poorly on some criteria and is of medium quality, the total score can be 5 to 6; \newline
When the quality of the model response is similar to the reference answer and performs well in all criteria, the total score should be 7 to 8; \newline
A score of 9 to 10 can only be achieved if the model significantly exceeds the quality of the reference answer, adequately addresses the user's question and all the needs, and is close to a perfect score on all criteria. 
As an example, the reference answer would receive a score of 8. \\ 
\bottomrule
    \end{tabular}
    \caption{English Instruction for evaluating Solve Professional Problem intent questions. Part 1.}
    \label{tab:English_instruction}
\end{table*}

\begin{table*}[t]
\centering
    \begin{tabular}{{p{1.5cm}p{13.5cm}}}
    \toprule
 Output \newline Format & You need to evaluate and explain before you score. Your explanation of each criterion needs to be followed by the scoring. After that, at the end of your answer, return all of your scores in the following dictionary format, including the curly brackets, and make sure that your scores are integers: \newline
\{'Dimension 1': scoring, 'Dimension 2': scoring, ... , 'Final Score': Score\}, e.g. \{'Factuality': 9, 'User Satisfaction': 6, ... , 'Final Score': 7\}. \\ \midrule
 Reference \newline Materials & User Question: ... \newline
 [Start of Reference Answer] ...... [End of reference answer] \newline
 [start of AI assistant's answer] ...... [end of AI assistant's answer]\\ \bottomrule
    \end{tabular}
    \caption{English Instruction for evaluating Solve Professional Problem intent questions. Part 2.}
    \label{tab:English_instruction_v2}
\end{table*}

\begin{table*}[t]
    \centering
    \begin{tabular}{{p{2.5cm}p{12.5cm}}}
    \toprule
     Intent-aware \newline Criteria    &  \chinese{请你以公正的评判者的身份，评估一个AI助手对于用户提问的回答的质量。由于您评估的回答类型是[解决专业问题]，因此你需要从下面的5个维度对回答进行评估: \newline
1 事实正确性(Factuality) \newline
提供的信息是否准确无误，是否基于可信的事实和数据  \newline
2 满足用户需求(User Satisfaction) \newline
是否满足了用户提出问题的目的和需求，是否对问题进行了全面而恰当的回应  \newline
3 清晰度(Clarity) \newline
是否表达清晰易懂，是否使用了简洁的语言和结构，以便用户可以轻松理解  \newline
4 逻辑连贯性(Logical Coherence) \newline
是否在整体上保持一致，是否在不同部分之间保持逻辑连贯性，避免了自相矛盾  \newline
5 完备性(Completeness) \newline
回答是否提供了足够的信息和细节，以满足用户的需求，是否遗漏了重要的方面 \newline
注意，回答不是越长越好，简短并且满足上述要求的回答是最好的。}\\ \midrule
    Chain of  \newline thought \newline reasoning steps     & \chinese{我们会给您提供用户的提问，一个8分左右的参考答案，和需要你评估的AI助手的答案。当你开始你的评估时，你需要按照遵守以下的流程：\newline
1. 将AI助手的答案与参考答案进行比较，指出AI助手的答案有哪些不足，并进一步解释。\newline
2. 从不同维度对AI助手的答案进行评价，在每个维度的评价之后，给每一个维度一个1～10的分数。\newline
3. 最后，综合每个维度的评估，对AI助手的回答给出一个1～10的综合分数。\newline
4. 你的打分需要尽可能严格，并且要遵守下面的评分规则：总的来说，模型回答的质量越高，则分数越高。\newline
其中，事实正确性和满足用户需求这两个维度是最重要的，这两个维度的分数主导了最后的综合分数。} \\ \midrule
Scoring \newline Standards & \chinese{当模型回答存在与问题不相关，或者有本质性的事实错误，或生成了有害内容时，总分必须是1到2分；\newline
当模型回答没有严重错误而且基本无害，但是质量较低，没有满足用户需求，总分为3到4分；\newline
当模型回答基本满足用户要求，但是在部分维度上表现较差，质量中等，总分可以得5到6分；\newline
当模型回答质量与参考答案相近，在所有维度上表现良好，总分得7到8分；\newline
只有当模型回答质量显著超过参考答案，充分地解决了用户问题和所有需求，并且在所有维度上都接近满分 的情况下，才能得9到10分。 
作为示例，参考答案可以得到8分。}\\ \midrule
 Output Format & \chinese{请记住，你必须在你打分前进行评价和解释。在你对每个维度的解释之后，需要加上对该维度的打分。之后，在你回答的末尾，按照以下字典格式（包括括号）返回你所有的打分结果，并确保你的打分结果是整数： \newline
\{’维度一’: 打分, ’维度二’: 打分, ..., ’综合得分’: 打分\}，例如：\{’事实正确性’: 9, ’满足用户需求’: 6, ...,’综合得分’: 7\}。}\\ \midrule
 Reference \newline Materials & \chinese{ 用户的提问：... \newline
 [参考答案开始] ...... [参考答案结束] \newline
 [助手的答案开始] ...... [助手的答案结束]}\\
\bottomrule
    \end{tabular}
    \caption{Chinese Instruction for evaluating Solve Professional Problem intent questions}
    \label{tab:chinese-instruction}
\end{table*}

\section{Benchmark Results}

\subsection{Bi-lingual Results}
\label{appendix:results_EN_CN}

Table~\ref{table:objective_results_ENCN} and Table~\ref{table:subjective_results_ENCN} show the English and Chinese's separated benchmark results across different intent.

\begin{table*}[t]
% \resizebox*{\textwidth}{!}{
\begin{tabular}{lcccccccc}
\toprule
                      & \multicolumn{2}{l}{Factual QA} & \multicolumn{2}{l}{Solve Problem} & \multicolumn{2}{l}{Text   Assistant}  & English & Chinese \\ \midrule
GPT-4o          & 8.55 & 8.47 & 8.42 & 8.21 & 8.11 & 7.64 & 8.40 & 8.26 \\
Qwen-max        & 8.30 & 7.44 & 8.14 & 7.44 & 8.38 & 8.12 & 8.26 & 7.49 \\
Claude-3-opus   & 8.05 & 7.64 & 8.12 & 7.54 & 7.62 & 7.69 & 7.97 & 7.58 \\
GLM-4           & 8.08 & 7.09 & 7.93 & 7.49 & 7.76 & 7.62 & 7.95 & 7.36 \\
Moonshot-v1-8k  & 7.67 & 7.56 & 8.00 & 7.22 & 7.57 & 7.95 & 7.77 & 7.39 \\
ERNIE-Bot-4     & 7.22 & 7.15 & 7.96 & 7.47 & 7.03 & 7.55 & 7.45 & 7.37 \\
GPT-3.5-turbo   & 7.45 & 6.55 & 7.07 & 6.48 & 7.35 & 6.79 & 7.29 & 6.52 \\
Baichuan2-Turbo & 7.33 & 6.67 & 6.68 & 6.57 & 6.57 & 7.29 & 6.91 & 6.65 \\
Deepseek-chat   & 6.75 & 6.07 & 6.86 & 6.70 & 6.78 & 6.83 & 6.80 & 6.49 \\
Spark-3.5       & 6.43 & 6.79 & 6.65 & 7.03 & 7.08 & 7.93 & 6.67 & 7.01\\ \bottomrule
\end{tabular} %}
\caption{Bi-lingual Benchmark Results. Part 1.}
\label{table:objective_results_ENCN}
\end{table*}

\begin{table*}[]
\begin{tabular}{lcccccccc}
\toprule
               & \multicolumn{2}{l}{Ask   for Advice} & \multicolumn{2}{l}{Seek   Creativity} & \multicolumn{2}{l}{Leisure} & English & Chinese \\ \midrule
GPT-4o          & 7.48 & 7.94 & 8.50 & 7.70 & 7.43 & 7.67 & 7.80 & 7.78 \\
GLM-4           & 7.45 & 7.14 & 8.07 & 6.93 & 7.50 & 6.25 & 7.66 & 6.80 \\
Qwen-max        & 7.59 & 7.21 & 7.64 & 6.99 & 7.88 & 6.49 & 7.66 & 6.92 \\
Moonshot-v1-8k  & 7.18 & 7.06 & 6.71 & 7.11 & 7.38 & 6.97 & 7.07 & 7.05 \\
Claude-3-opus   & 6.86 & 7.04 & 7.21 & 7.06 & 7.13 & 7.16 & 7.02 & 7.08 \\
GPT-3.5-turbo   & 6.64 & 6.29 & 6.79 & 6.13 & 7.38 & 5.57 & 6.82 & 6.02 \\
Deepseek-chat   & 6.55 & 5.98 & 7.00 & 5.27 & 7.13 & 4.69 & 6.80 & 5.37 \\
ERNIE-Bot-4     & 6.95 & 7.13 & 6.64 & 7.30 & 6.25 & 7.11 & 6.73 & 7.17 \\
Baichuan2-Turbo & 6.27 & 6.37 & 6.64 & 6.00 & 7.00 & 5.92 & 6.52 & 6.12 \\
Spark-3.5       & 6.43 & 6.79 & 6.65 & 7.03 & 7.08 & 7.93 & 6.67 & 7.01\\    \bottomrule
\end{tabular}
\caption{Bi-lingual Benchmark Results. Part 2.}
\label{table:subjective_results_ENCN}
\end{table*}

\begin{table}[]
    \centering
    \begin{tabular}{lr}
    \toprule
    Service & Score \\ \midrule
    GPT-4  & 8.63\\
    GPT-3.5 & 6.95 \\ \hline
    Search Engine (Google Top-1 page)     & 5.14 \\
    RAG (GPT-4 + Google Top-5 page)    & 6.47 \\ \bottomrule
    \end{tabular}
    \caption{Comparison between LLM services and Search Engine in 57 Factual QA cases from the URS dataset.}
    \label{table:search_engine}
\vspace{-2ex}
\end{table}

\begin{table}[]
    \centering
    \begin{tabular}{lr}
    \toprule
    Service & Score \\ \midrule
    GPT-4  & 9.00\\
    GPT-3.5 & 8.50  \\ \hline
    Translator (Google)    & 7.00 \\\bottomrule
    \end{tabular}
    \caption{Comparative Evaluation of LLM Services and Translator in 8 translation-related Text Assistant Scenarios from the URS dataset.}
    \label{table:translator}
\vspace{-2ex}
\end{table}

\subsection{Scoring Examples}
\label{appendix:scoring_example}
For subjective intents, we list the example cases as shown in Table~\ref{tab:evaluation_cases}, \ref{tab:evaluation_cases_v2} and \ref{tab:evaluation_cases_v3}.
% \todo{cases}
% False Premises
% Creativity, Humor,
% implicit user intent
% cultural background

\subsection{Alignment with Human Annotation}
\label{appendix:alignment}

\begin{figure}[t]
    \centering
    \includegraphics[width=0.9\linewidth]{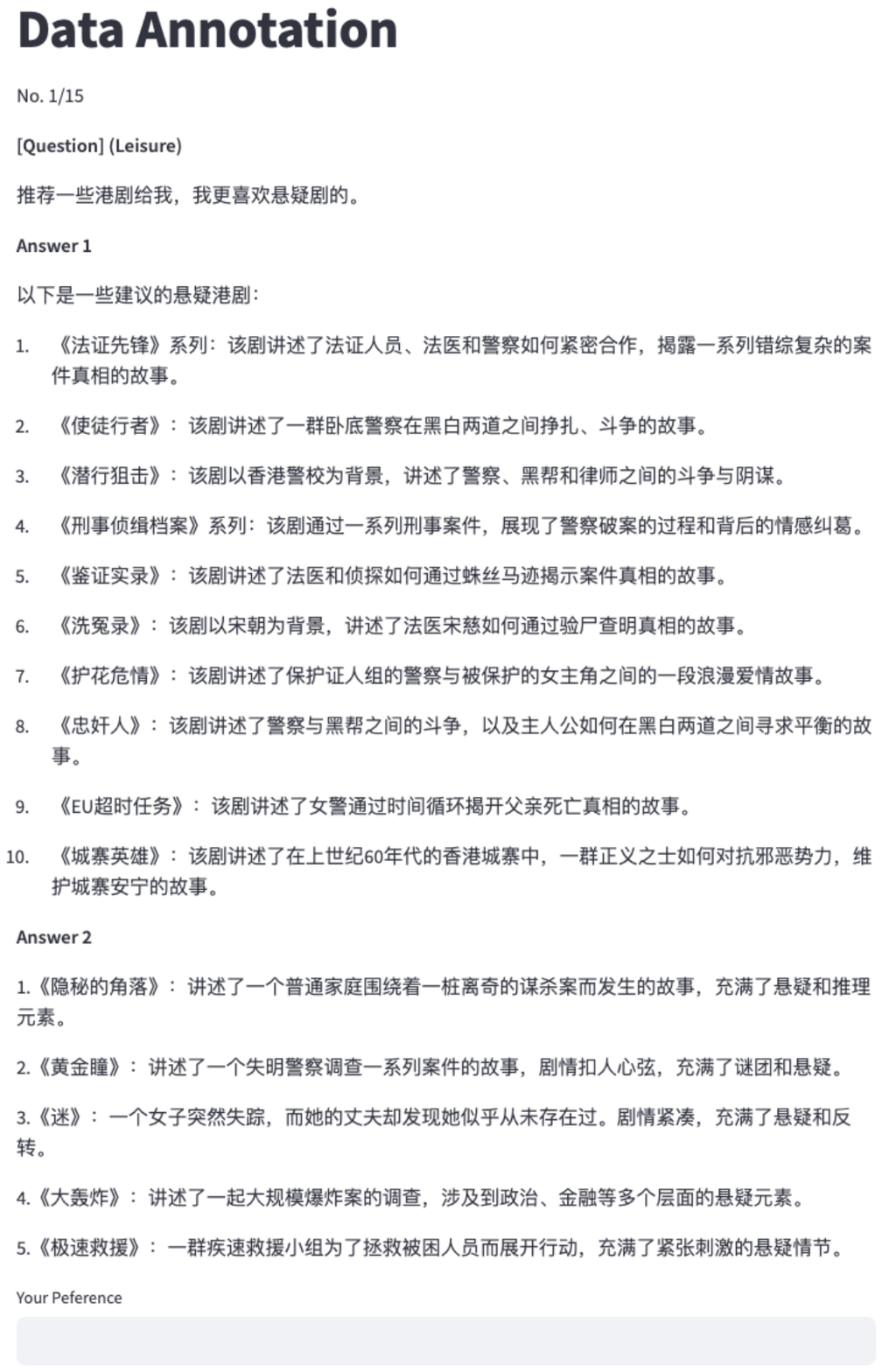}
    \caption{Data Annotation Page.}
    \label{fig:data_annotation}
\end{figure}

We aim to further investigate whether the benchmarked LLM order aligns with human preferences by conducting annotation experiments. Following the method outlined in Chatbot Arena~\cite{chiang2024chatbot}, we enlisted five human annotators (three males and two females) to evaluate paired questions as shown on the UI in Figure~\ref{fig:data_annotation}.
We randomly selected 400 questions from the 1024-case benchmark and randomly paired two LLM outputs for each question. Annotators, blind to which LLMs generated the responses, were tasked with choosing their preferred answer from the two provided, or indicating if the choice was equal or undeterminable due to their knowledge limitations. Of the 400 paired comparisons, 50 were marked "cannot be determined." Utilizing the remaining annotations, we calculated the Bradley-Terry (BT) scores to establish LLM rankings based on pairwise competition outcomes. As shown in Figure~\ref{figure:alignment_LLM}, the resulting LLM order aligns with the benchmark rankings, and the BT score demonstrates a Pearson correlation of 0.94 with the benchmark score.
The evaluation data is anonymously available.

% We want to further examine whether the benchmark LLM order aligns with human preference and conduct human annotation experiments.
% Following the evaluation method of Chatbot Arena~\cite{chiang2024chatbot}, we recruit 5 human annotators~(3 males and 2 females, aged from 20 to 30) to compare paired questions as illustrated in Figure~\ref{fig:data_annotation}.
% We randomly selected 400 questions from the 1024 cases benchmark and randomly selected two LLMs' outputs. Annotators were asked to compare the two answers under each question, without knowing which LLMs generated the answer, and select whether they preferred Answer 1, Answer 2, equal or cannot be determined due to their knowledge limitations. Among the 400 pair comparisons, 50 cases received "cannot be determined". Based on the 350 annotations, we compute the BT score to get LLM orders based on the pair competition situations. The LLM order aligns with the benchmark order and the BT-score has a Pearson correlation of 0.94 with the benchmark score.

\subsection{Comparison with Different Evaluation Method}
% In the overall benchmark result, we adopt GPT-4's answer as the reference for score 8 and use GPT-4 as the evaluator base for final scorings.
% This might cause biases towards answers similar to GPT. Thus, we re-evaluate with another 2 LLMs, Qwen and GLM, replacing the reference answer and base evaluator, respectively.

\begin{figure}[t]
    \centering
    \includegraphics[width=0.99\linewidth]{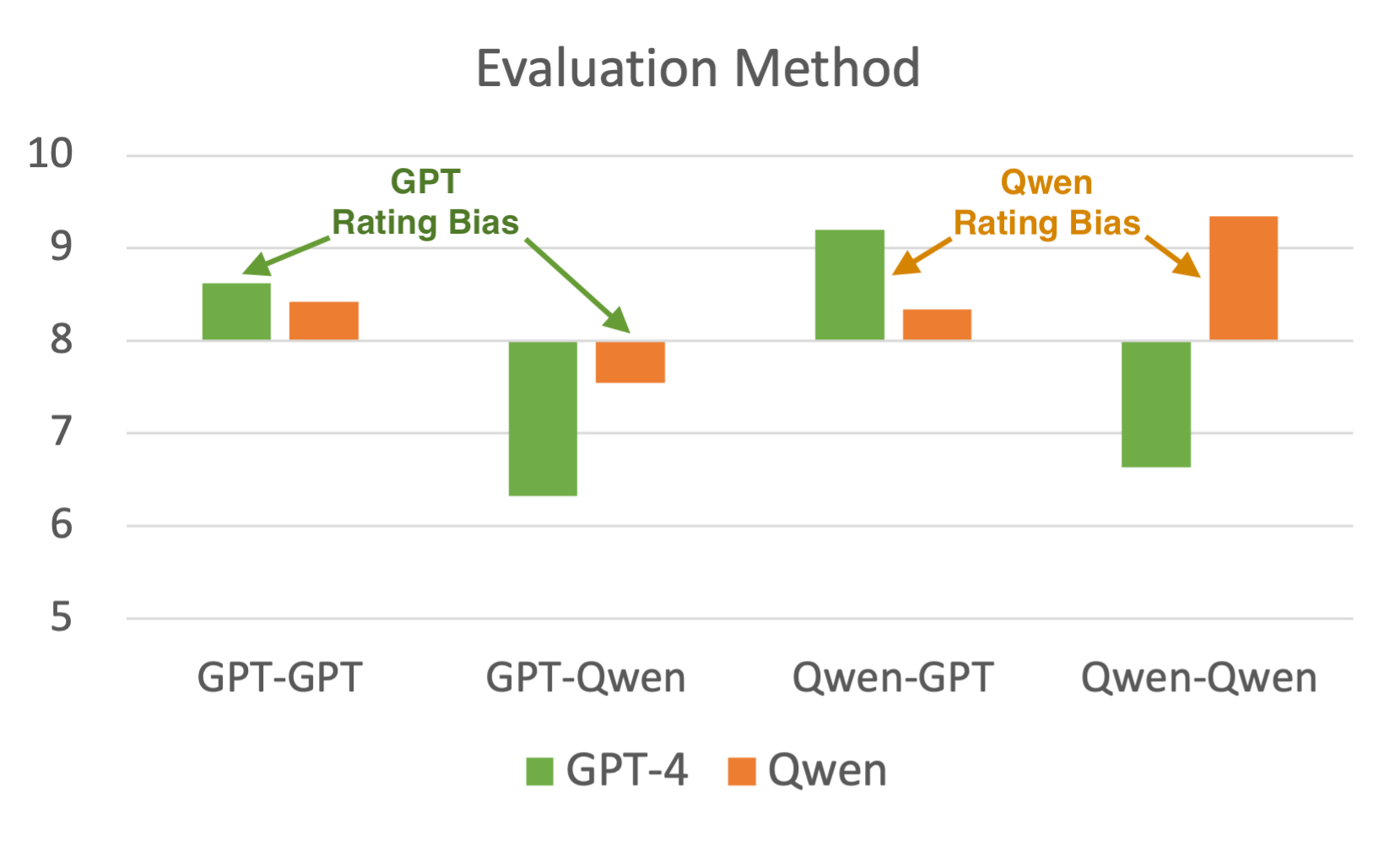}
    \includegraphics[width=0.99\linewidth]{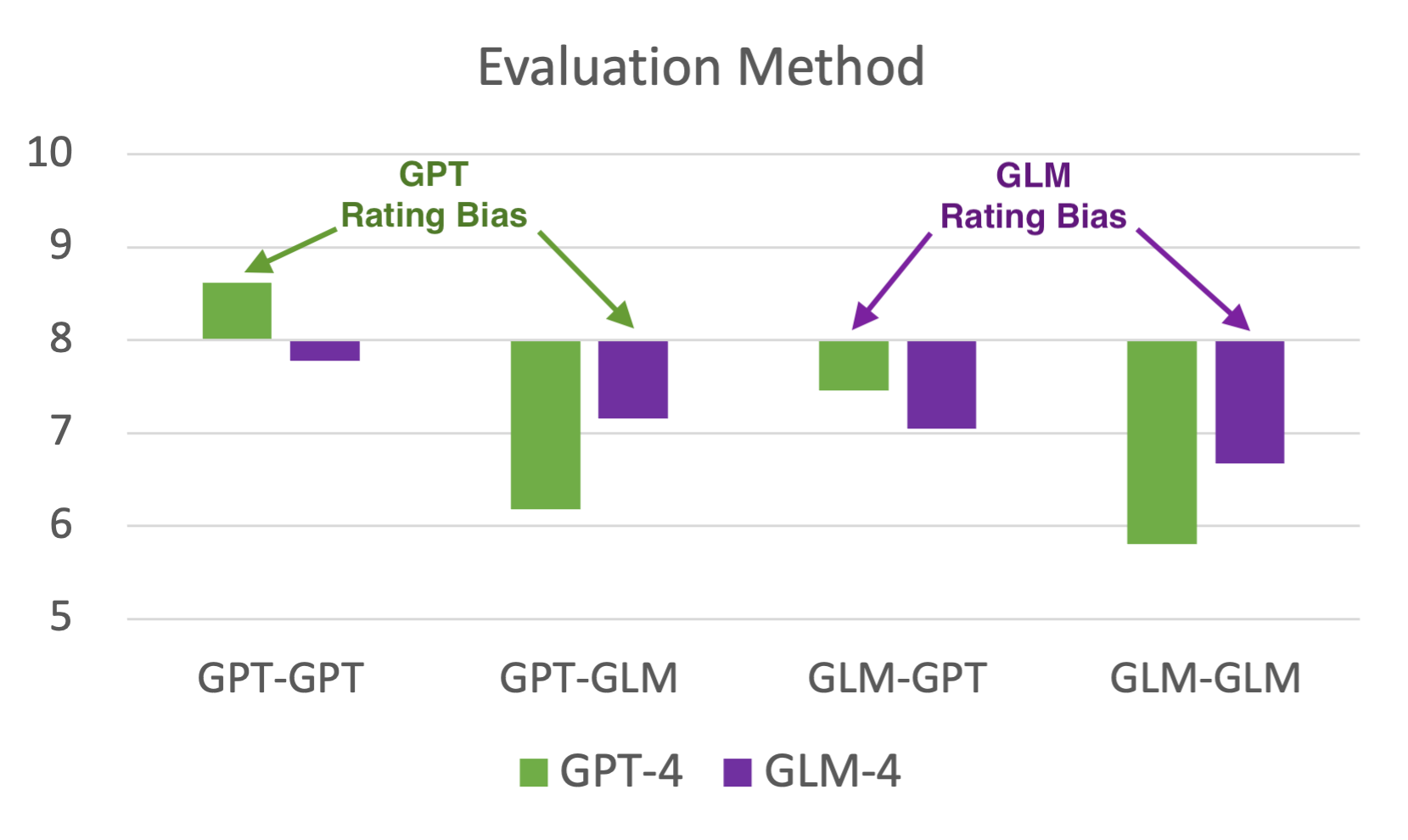}
    \caption{Analysis of Evaluation Methodologies: "A-B" indicates the evaluation setting, where 'A' denotes the evaluator LLM and 'B' represents the source LLM used to generate 8-point reference answers.
    Rating bias is defined in Equation~\ref{equ:rating_bias}. Specifically, the rating bias for evaluator A corresponds to the discrepancy between the reference score, 8, and the benchmark score of Model B under the assessment setting of A-B.}
    \label{fig:evaluation_method_v2}
\end{figure}

Figure~\ref{fig:evaluation_method_v2} shows cross-validation evaluation results. We change the model to generate reference answers and the model to serve as evaluators respectively. Results show while using GPT4 output as the reference answer for 8 scores, the order between the three models remains unchanged no matter the evaluation model.
While using Qwen or GLM's output as both reference generator and evaluator, there exists a strong incline towards the model itself.

% \todo{not finish. more group of resutls}

% \todo{format wrong rate show}

We also check the rating bias of different LLMs as evaluators. Rating bias is defined in Equation~\ref{equ:rating_bias}.
\begin{equation}
  \text{Rating Bias}(A)=\frac{\sum_{B}\text{Rating}_{A-B}(B)-r}{\sum_B 1},\\
  r=8
  \label{equ:rating_bias}
\end{equation}
where $\text{Rating}_{A-B}(B)$ representing using LLM A as an evaluator and the output of LLM B as a reference answer to evaluate the performance of LLM B. $r$ is the score for reference, in our cases, it is equal to 8.
Note that in the experiments, reference answer and output awaiting score are two-time outputs of the same model, so if model B is stable and model A can score accurately, the $\text{Rating}_{A-B}(B)$ should be $r$. We measure the $\text{Rating Bias}$ to measure the calibrability of the evaluator model.
As shown in the results, compared to Qwen-max and GLM-4, GPT-4 performs more accurately in this cross-validation experiments.

% Results are shown in Figure~\ref{fig:evaluation_method_v2}.
\subsection{Case Study}
% \subsubsection{Computational Capabilities}
In our benchmark, we incorporate a diverse range of scenarios, extending beyond commonplace applications to include highly specialized use cases.
% particularly in the intent of API utilization. 
A notable example involves assessing the models' proficiency in executing sequential summations without relying on external tools. This particular computational skill often manifests spontaneously and poses a challenge for targeted training. Consequently, evaluating a model's capability to accurately perform a series of two-digit additions offers a rapid and effective indicator of its computational prowess.

Our examination entails the concatenation of two-digit numbers ranging from 2 to 20 in sequence, with each magnitude category being subjected to 10 randomly generated test instances. Table~\ref{table:addition} delineates the performances of the top LLMs, showing their ability to achieve the longest chain of consecutive additions while maintaining a minimum accuracy threshold of 80\% and the accuracy of adding from 2 to 10 numbers. Models not listed achieved an accuracy lower than 50\%.

% These findings demonstrate a correlation with the models' scores in the "Solve Professional Problem" intent within our benchmark. The simplicity and efficiency of this testing methodology render it an advantageous approach for swiftly gauging the computational capacity of large language models.

It is worth noting that the top 4 LLMs with maximum length$\textgreater$5 are the ones with larger parameters out of the total 10 LLMs. This might indicate that the ability of computing has correlations with the scaling of model size.

\subsection{Scenario Study}
Since we evaluate LLMs as collaborative tools for users, we juxtapose their performance with traditional assistants like search engines and translators. This comparison can assist and guide users in choosing the most effective tool for their specific intents.
To this end, we evaluate LLM performance compared to Google Search Engine for Factual QA intent questions and Google Translator for translation queries in Text Assistant scenarios.

\noindent \textbf{Search Engine.}
%We utilize the SerpAPI package\footnote{https://serpapi.com/}, a Google Search API, to retrieve search results for 57 random questions within the Factual QA intent.
We first retrieve search results for 57 random questions within the Factual QA intent and
devise two methods to form answers: 1) Direct search results: the content of the top-1 web page, and 2) Retrieval-augmented generation: GPT-4 summarized answer based solely on information from the top-5 pages. Note that this approach enhances the search performances, not the LLMs'.
The results are reported in Table~\ref{table:search_engine}.
Our analysis shows that direct search results do not outperform GPT-3.5. This limitation can be attributed to the fact that web pages might not provide straightforward answers to user queries, requiring users to navigate through multiple pages to piece together a response. The retrieval-augmented generation approach, which employs GPT-4 to create summarized answers based on search results without incorporating its inherent knowledge, also underperforms LLMs.
% We observe that the direct search results cannot perform better than GPT-3.5.
% This can be attributed to the fact that web pages do not always present direct answers to user queries, necessitating users to sift through multiple pages to formulate an answer.
% The retrieval-augmented generation used GPT-4 to create summarized answers based on the search page content without incorporating its inherent knowledge base. Note that this method here is an augmentation for search, which is not an augmentation to LLMs.
% This method also performs worse than GPT-3.5.
These observations underscore LLMs' efficacy in factual QA scenarios, likely due to their ability to respond directly to user queries. 
% The fact that RAG underperforms LLMs for this task suggests that LLMs have not effectively incorporated the search results in the generation of answers.
% We can see that the direct page fails behind the bottom LLM results. This is partly due to web pages sometimes do not directly provide the exact answer to the question. While using search engines, we need to go through several pages in order to form the answer by our own. Thus, we test the retrieval augmented generation setting, which provide the question, search page links and ask GPT-4 to summarized answers without usage of this own knowledge. This results still not pass the GPT-3.5, never to mention GPT-4. This show that LLMs performance well in the factual QA scenario this might because it can provide direct answer to user queries.

% False Premises
% Creativity, Humor,
% implicit user intent
% cultural background

\noindent \textbf{Translation.}
%In text assistant scenarios, users have options of using translators or LLM services to fulfill their translation needs.
%We select the translation cases in our URS dataset, and evaluate the Google Translator\footnote{https://translate.google.com} output.
To simulate the text assistant scenarios where users have several options of choosing translators or LLM services for their translation needs, we select the translation cases in our URS dataset for evaluation.
Results are shown in Table~\ref{table:translator}, where we find LLM services outperform the translation tool. 
This may be because users need specific styles or modifications of the original context, while traditional translators are hard to accomplish.

\begin{table*}[t]
    \centering
    \begin{tabular}{{p{2cm}p{12cm}p{1cm}}}
    \toprule
     Intent\newline Question &Seek Creativity\newline \chinese{如何看待ai将会导致大部分人失业的观点}  &  \\ \midrule
     Reference\newline Answer & \chinese{AI的发展确实可能会对一些职业产生影响，导致一部分工作被机器取代。但这并不意味着会导致大部分人失业，主要有以下几点原因。

1. AI也会创造新的就业机会。尽管AI可能会替代一些人力密集型的工作，但同时也有可能创造出新的工作岗位。例如，现在已经有很多公司开始雇佣AI相关的技术人员，如数据科学家、机器学习工程师等。

2. AI可能会提高人们的工作效率，从而让人们有更多的时间投入到其他需要人类特质(如创造性，同理心)的工作中去。

3. 社会和政府也可以采取一些政策，如教育改革、职业培训、调整税收政策等，以帮助人们适应新的就业形式。

总的来说，AI的发展可能会改变就业市场的面貌，但不一定会导致大部分人失业。我们需要做的是积极应对，持续学习，提升自己的竞争力。}  &\\ \midrule
     Claude-3 & \chinese{关于AI可能导致大规模失业的观点,我的看法是:

1. AI技术的发展和应用确实会在一定程度上改变就业市场,一些工作岗位可能会被AI取代,尤其是那些高度重复、程式化的工作。但同时AI也会创造出许多新的就业机会。

2. 从历史来看,每一次技术革命都伴随着部分职业的消失和新职业的出现。比如工业革命导致手工业者失业,但创造了更多工厂工人的岗位。因此AI带来的是就业结构的转型和升级,而不是单纯的失业。

3. 为了适应AI时代,教育和职业培训需要与时俱进。学校应加强STEM教育,培养学生的科技素养;企业和政府应积极开展职业培训,帮助员工掌握新技能。终身学习将成为常态。

4. 政府在AI时代扮演着至关重要的角色。一方面要大力发展AI产业,抢占技术制高点;另一方面要完善社会保障体系,为就业转型提供安全网,尽量降低阵痛。

5. AI的最终目标应该是用技术造福人类,而不是替代人类。在就业问题上,我们要努力实现人机协作,发挥人工智能和人类智慧的各自优势,创造更加智能和人性化的工作环境。

总之,对待AI可能带来的失业风险,我们需要保持乐观和开放的心态,未雨绸缪,在顺应技术发展趋势的同时,努力创造一个更加包容和可持续的未来。这需要全社会共同努力。}  & Final \newline Score:\newline 8\\ 
\bottomrule
    \end{tabular}
    \caption{Evaluation Cases}
    \label{tab:evaluation_cases}
\end{table*}

\begin{table*}[t]
    \centering
    \begin{tabular}{{p{2cm}p{12cm}p{1cm}}}
    \toprule
     Intent\newline Question &Seek Creativity\newline \chinese{有没有一些能3d打印的点子}  & \\ \midrule
     Reference\newline Answer & \chinese{1. 个性化首饰：使用3D打印技术可以打造出一些独一无二的首饰款式，如戒指、项链、耳环等。

2. 手机壳：可以按照手机的具体型号和个人喜好设计特色手机壳。

3. 家居装饰品：如烛台、相框、花盆等。

4. 动手玩具：例如益智拼图，也可以设计特色的棋盘游戏。

5. 办公室工具：如笔筒、名片架等。

6. 厨房用具：例如饼干模具，也可以设计独特的餐具，如喝咖啡的杯子、调料瓶等。

7. 个性化口罩：在疫情期间，可以设计一些有趣的3D口罩。

8. 最小化家具模型：这对于室内设计师来说非常有用，他们可以创建一个房间的精确比例模型。

9. 自定义剃须刀把手：通过3D打印，你可以轻松获得符合你个人喜好和握手方式的剃须刀把手。

10. 皮肤模型：对皮肤科医生，可以打印出一些皮肤疾病的三维模型，有助于他们解释疾病的具体情况。 

11. 3D打印的鞋子：通过精确测量脚的尺寸，可以创建完全符合脚型的鞋子。

要注意的是有些物品例如食品、口罩等需要考虑材料的安全性和舒适性。在3D打印任何物品时，一定要注意材料的选择和劳动安全。}  &\\ \midrule
     Claude-3 & \chinese{当然有很多适合3D打印的创意点子,这里列举一些供参考:

1. 个性化手机壳或支架 - 可以根据自己的喜好设计独特的手机配件。

2. 定制化礼物 - 如姓名钥匙扣、纪念品等,送给亲朋好友很有心意。

3. 桌面小摆设 - 打印一些有创意的小装饰品点缀桌面。

4. 益智玩具或拼图 - 打印各种有趣的玩具,锻炼动手和思维能力。

5. 文具小物件 - 如笔筒、尺子、铅笔套等学习用品。

6. 厨房用具 - 打印些创意的饼干模具、蛋糕装饰物等。

7. 宠物用品 - 定制宠物饮食碗、玩具等,让爱宠与众不同。

8. Cosplay道具 - 打印出各种动漫人物的武器、面具等。

9. 家居生活用品 - 如衣帽钩、牙刷架、杯垫等实用物件。

10. 自制简单机械 - 打印一些小发明,如各种齿轮组、简易机器人等。

11. 园艺工具 - 如花盆、铲子、花洒喷头等。

12. 艺术品或雕塑 - 将你的创意艺术构想变为现实。

总之3D打印的应用非常广泛,发挥你的创意,定制专属自己的物品吧。建议在打印前先做好建模设计,以保证成品质量。}  & Final\newline Score: \newline 7\\ 
\bottomrule
    \end{tabular}
    \caption{Evaluation Cases}
    \label{tab:evaluation_cases_v2}
\end{table*}

\begin{table*}[t]
    \centering
    \begin{tabular}{{p{2cm}p{12cm}p{1cm}}}
    \toprule
     Intent\newline Question &Seek Creativity\newline \chinese{日常补充剂的推荐摄入量是多少？}  &  \\ \midrule
     Reference\newline Answer & \chinese{推荐的日常补充剂摄入量因年龄、性别、健康状况和特定需求而异。以下是几种常见补充剂的大致指导量：

1. **维生素D**: 成人每日推荐400-800 IU（国际单位），但许多专家推荐每日1000-2000 IU以优化血液中的维生素D水平。

2. **Omega-3脂肪酸（如鱼油）**: 对于心脏健康，推荐每日摄入EPA和DHA的总量为250-500mg。

3. **维生素C**: 成人推荐每日摄入量为男性90mg，女性75mg。

4. **维生素B12**: 成人推荐每日摄入量为2.4μg。

5. **钙**: 成人建议每日摄入量为1000mg，50岁以上的女性和70岁以上的男性建议每日摄入量增至1200mg。

6. **铁**: 成人男性推荐每日摄入量为8mg，成年女性建议18mg（根据月经周期可能需要调整）。

7. **镁**: 成人男性建议每日摄入量为400-420mg，女性为310-320mg。

8. **锌**: 成人男性每日推荐11mg，女性推荐8mg。

重要提示：这些数量只是大致指导，个人需求可能有所不同。服用任何补充剂前，建议咨询医生或营养专家，以获得个人化的建议和避免超量摄入。超量摄入某些营养素可能会对健康产生负面影响。}  &\\ \midrule
     Qwen-max & \chinese{补充剂的推荐摄入量因具体的补充剂种类和个体差异而异，通常无法一概而论。例如：

1. 维生素C：成年人每日推荐摄入量为75-90毫克，但上限一般不超过2000毫克。

2. 钙：成年人每日推荐摄入量约为800-1000毫克，老年人可能需要更多。

3. 维生素D：成年人推荐摄入量一般为每天400-800国际单位，但某些专家建议更高，如每天1000-2000国际单位。

在考虑补充任何营养素之前，最好是先咨询医生或营养师的意见，他们可以根据你的年龄、性别、健康状况、饮食习惯以及是否存在特殊需求（如孕期、哺乳期、疾病恢复期等）来给出个性化的建议。同时，需要注意的是，尽管补充剂可以弥补饮食中某些营养素的不足，但合理膳食始终是获取充足营养的首选方式。过量摄入某些补充剂可能会对健康造成负面影响。}  & Final \newline Score:\newline 5\\ 
\bottomrule
    \end{tabular}
    \caption{Evaluation Cases}
    \label{tab:evaluation_cases_v3}
\end{table*}

%% file: acl2023.bbl
\begin{thebibliography}{47}
\expandafter\ifx\csname natexlab\endcsname\relax\def\natexlab#1{#1}\fi

\bibitem[{Achiam et~al.(2023)Achiam, Adler, Agarwal, Ahmad, Akkaya, Aleman, Almeida, Altenschmidt, Altman, Anadkat et~al.}]{ChatGPT}
Josh Achiam, Steven Adler, Sandhini Agarwal, Lama Ahmad, Ilge Akkaya, Florencia~Leoni Aleman, Diogo Almeida, Janko Altenschmidt, Sam Altman, Shyamal Anadkat, et~al. 2023.
\newblock Gpt-4 technical report.
\newblock \emph{arXiv preprint arXiv:2303.08774}.

\bibitem[{Anthropic(2024)}]{Claude}
AI~Anthropic. 2024.
\newblock The claude 3 model family: Opus, sonnet, haiku.
\newblock \emph{Claude-3 Model Card}.

\bibitem[{Austin et~al.(2021)Austin, Odena, Nye, Bosma, Michalewski, Dohan, Jiang, Cai, Terry, Le, and Sutton}]{austin2021program}
Jacob Austin, Augustus Odena, Maxwell Nye, Maarten Bosma, Henryk Michalewski, David Dohan, Ellen Jiang, Carrie Cai, Michael Terry, Quoc Le, and Charles Sutton. 2021.
\newblock \href {http://arxiv.org/abs/2108.07732} {Program synthesis with large language models}.

\bibitem[{Bai et~al.(2024)Bai, Liu, Bu, He, Liu, Zhou, Lin, Su, Ge, Zheng et~al.}]{bai2024mt}
Ge~Bai, Jie Liu, Xingyuan Bu, Yancheng He, Jiaheng Liu, Zhanhui Zhou, Zhuoran Lin, Wenbo Su, Tiezheng Ge, Bo~Zheng, et~al. 2024.
\newblock Mt-bench-101: A fine-grained benchmark for evaluating large language models in multi-turn dialogues.
\newblock \emph{arXiv preprint arXiv:2402.14762}.

\bibitem[{Bai et~al.(2023)Bai, Bai, Chu, Cui, Dang, Deng, Fan, Ge, Han, Huang et~al.}]{Qwen}
Jinze Bai, Shuai Bai, Yunfei Chu, Zeyu Cui, Kai Dang, Xiaodong Deng, Yang Fan, Wenbin Ge, Yu~Han, Fei Huang, et~al. 2023.
\newblock Qwen technical report.
\newblock \emph{arXiv preprint arXiv:2309.16609}.

\bibitem[{Baidu()}]{ERNIE}
Baidu.
\newblock \href {https://yiyan.baidu.com/} {Ernie bot}.

\bibitem[{bench authors(2023)}]{srivastava2023beyond}
BIG bench authors. 2023.
\newblock \href {https://openreview.net/forum?id=uyTL5Bvosj} {Beyond the imitation game: Quantifying and extrapolating the capabilities of language models}.
\newblock \emph{Transactions on Machine Learning Research}.

\bibitem[{Bi et~al.(2024)Bi, Chen, Chen, Chen, Dai, Deng, Ding, Dong, Du, Fu et~al.}]{bi2024deepseek}
Xiao Bi, Deli Chen, Guanting Chen, Shanhuang Chen, Damai Dai, Chengqi Deng, Honghui Ding, Kai Dong, Qiushi Du, Zhe Fu, et~al. 2024.
\newblock Deepseek llm: Scaling open-source language models with longtermism.
\newblock \emph{arXiv preprint arXiv:2401.02954}.

\bibitem[{Bodonhelyi et~al.(2024)Bodonhelyi, Bozkir, Yang, Kasneci, and Kasneci}]{bodonhelyi2024user}
Anna Bodonhelyi, Efe Bozkir, Shuo Yang, Enkelejda Kasneci, and Gjergji Kasneci. 2024.
\newblock User intent recognition and satisfaction with large language models: A user study with chatgpt.
\newblock \emph{arXiv preprint arXiv:2402.02136}.

\bibitem[{Bolotova et~al.(2022)Bolotova, Blinov, Scholer, Croft, and Sanderson}]{bolotova2022non}
Valeriia Bolotova, Vladislav Blinov, Falk Scholer, W~Bruce Croft, and Mark Sanderson. 2022.
\newblock A non-factoid question-answering taxonomy.
\newblock In \emph{Proceedings of the 45th International ACM SIGIR Conference on Research and Development in Information Retrieval}, pages 1196--1207.

\bibitem[{Bradley and Terry(1952)}]{bradley1952rank}
Ralph~Allan Bradley and Milton~E Terry. 1952.
\newblock Rank analysis of incomplete block designs: I. the method of paired comparisons.
\newblock \emph{Biometrika}, 39:324--345.

\bibitem[{Chang et~al.(2023)Chang, Wang, Wang, Wu, Yang, Zhu, Chen, Yi, Wang, Wang et~al.}]{chang2023survey}
Yupeng Chang, Xu~Wang, Jindong Wang, Yuan Wu, Linyi Yang, Kaijie Zhu, Hao Chen, Xiaoyuan Yi, Cunxiang Wang, Yidong Wang, et~al. 2023.
\newblock A survey on evaluation of large language models.
\newblock \emph{ACM Transactions on Intelligent Systems and Technology}.

\bibitem[{Chen et~al.(2021)Chen, Tworek, Jun, Yuan, Pinto, Kaplan, Edwards, Burda, Joseph, Brockman et~al.}]{chen2021evaluating}
Mark Chen, Jerry Tworek, Heewoo Jun, Qiming Yuan, Henrique Ponde de~Oliveira Pinto, Jared Kaplan, Harri Edwards, Yuri Burda, Nicholas Joseph, Greg Brockman, et~al. 2021.
\newblock Evaluating large language models trained on code.
\newblock \emph{arXiv preprint arXiv:2107.03374}.

\bibitem[{Chiang et~al.(2024)Chiang, Zheng, Sheng, Angelopoulos, Li, Li, Zhang, Zhu et~al.}]{chiang2024chatbot}
Wei-Lin Chiang, Lianmin Zheng, Ying Sheng, Anastasios~Nikolas Angelopoulos, Tianle Li, Dacheng Li, Hao Zhang, Banghua Zhu, et~al. 2024.
\newblock Chatbot arena: An open platform for evaluating llms by human preference.
\newblock \emph{arXiv preprint arXiv:2403.04132}.

\bibitem[{Chu et~al.(2024)Chu, Ai, Tu, Li, and Liu}]{chu2024pre}
Zhumin Chu, Qingyao Ai, Yiteng Tu, Haitao Li, and Yiqun Liu. 2024.
\newblock Pre: A peer review based large language model evaluator.
\newblock \emph{arXiv preprint arXiv:2401.15641}.

\bibitem[{Clark et~al.(2018)Clark, Cowhey, Etzioni, Khot, Sabharwal, Schoenick, and Tafjord}]{clark2018think}
Peter Clark, Isaac Cowhey, Oren Etzioni, Tushar Khot, Ashish Sabharwal, Carissa Schoenick, and Oyvind Tafjord. 2018.
\newblock Think you have solved question answering? try arc, the ai2 reasoning challenge.
\newblock \emph{arXiv preprint arXiv:1803.05457}.

\bibitem[{Cobbe et~al.(2021)Cobbe, Kosaraju, Bavarian, Chen, Jun, Kaiser, Plappert, Tworek, Hilton, Nakano, Hesse, and Schulman}]{cobbe2021gsm8k}
Karl Cobbe, Vineet Kosaraju, Mohammad Bavarian, Mark Chen, Heewoo Jun, Lukasz Kaiser, Matthias Plappert, Jerry Tworek, Jacob Hilton, Reiichiro Nakano, Christopher Hesse, and John Schulman. 2021.
\newblock Training verifiers to solve math word problems.
\newblock \emph{arXiv preprint arXiv:2110.14168}.

\bibitem[{Du et~al.(2021)Du, Qian, Liu, Ding, Qiu, Yang, and Tang}]{du2021glm}
Zhengxiao Du, Yujie Qian, Xiao Liu, Ming Ding, Jiezhong Qiu, Zhilin Yang, and Jie Tang. 2021.
\newblock Glm: General language model pretraining with autoregressive blank infilling.
\newblock \emph{arXiv preprint arXiv:2103.10360}.

\bibitem[{Elangovan et~al.(2024)Elangovan, Liu, Xu, Bodapati, and Roth}]{elangovan2024considers}
Aparna Elangovan, Ling Liu, Lei Xu, Sravan Bodapati, and Dan Roth. 2024.
\newblock Considers-the-human evaluation framework: Rethinking human evaluation for generative large language models.
\newblock \emph{ACL 2024}.

\bibitem[{Feldhus et~al.(2023)Feldhus, Wang, Anikina, Chopra, Oguz, and M{\"o}ller}]{feldhus2023interrolang}
Nils Feldhus, Qianli Wang, Tatiana Anikina, Sahil Chopra, Cennet Oguz, and Sebastian M{\"o}ller. 2023.
\newblock Interrolang: Exploring nlp models and datasets through dialogue-based explanations.
\newblock \emph{arXiv preprint arXiv:2310.05592}.

\bibitem[{He et~al.(2024{\natexlab{a}})He, Luo, Bai, Hu, Thai, Shen, Hu, Han, Huang, Zhang et~al.}]{he2024olympiadbench}
Chaoqun He, Renjie Luo, Yuzhuo Bai, Shengding Hu, Zhen~Leng Thai, Junhao Shen, Jinyi Hu, Xu~Han, Yujie Huang, Yuxiang Zhang, et~al. 2024{\natexlab{a}}.
\newblock Olympiadbench: A challenging benchmark for promoting agi with olympiad-level bilingual multimodal scientific problems.
\newblock \emph{arXiv preprint arXiv:2402.14008}.

\bibitem[{He et~al.(2024{\natexlab{b}})He, Luo, Hu, Zhao, Zhou, Wu, Zhang, Han, Liu, and Sun}]{he2024ultraeval}
Chaoqun He, Renjie Luo, Shengding Hu, Yuanqian Zhao, Jie Zhou, Hanghao Wu, Jiajie Zhang, Xu~Han, Zhiyuan Liu, and Maosong Sun. 2024{\natexlab{b}}.
\newblock Ultraeval: A lightweight platform for flexible and comprehensive evaluation for llms.
\newblock \emph{ACL 2024 demo paper}.

\bibitem[{Hendrycks et~al.(2020)Hendrycks, Burns, Basart, Zou, Mazeika, Song, and Steinhardt}]{hendrycks2020measuring}
Dan Hendrycks, Collin Burns, Steven Basart, Andy Zou, Mantas Mazeika, Dawn Song, and Jacob Steinhardt. 2020.
\newblock Measuring massive multitask language understanding.
\newblock \emph{arXiv preprint arXiv:2009.03300}.

\bibitem[{Hendrycks et~al.(2021)Hendrycks, Burns, Kadavath, Arora, Basart, Tang, Song, and Steinhardt}]{hendrycks2021measuring}
Dan Hendrycks, Collin Burns, Saurav Kadavath, Akul Arora, Steven Basart, Eric Tang, Dawn Song, and Jacob Steinhardt. 2021.
\newblock Measuring mathematical problem solving with the math dataset.
\newblock \emph{arXiv preprint arXiv:2103.03874}.

\bibitem[{Hershcovich et~al.(2022)Hershcovich, Frank, Lent, de~Lhoneux, Abdou, Brandl, Bugliarello, Piqueras, Chalkidis et~al.}]{hershcovich2022challenges}
Daniel Hershcovich, Stella Frank, Heather Lent, Miryam de~Lhoneux, Mostafa Abdou, Stephanie Brandl, Emanuele Bugliarello, Laura~Cabello Piqueras, Ilias Chalkidis, et~al. 2022.
\newblock Challenges and strategies in cross-cultural nlp.
\newblock \emph{ACL 2022}.

\bibitem[{Huang et~al.(2024{\natexlab{a}})Huang, Qu, Liu, Yang, and Zhao}]{huang2024empirical}
Hui Huang, Yingqi Qu, Jing Liu, Muyun Yang, and Tiejun Zhao. 2024{\natexlab{a}}.
\newblock An empirical study of llm-as-a-judge for llm evaluation: Fine-tuned judge models are task-specific classifiers.
\newblock \emph{arXiv preprint arXiv:2403.02839}.

\bibitem[{Huang et~al.(2024{\natexlab{b}})Huang, Bai, Zhu, Zhang, Zhang, Su, Liu, Lv, Zhang, Fu et~al.}]{huang2024c}
Yuzhen Huang, Yuzhuo Bai, Zhihao Zhu, Junlei Zhang, Jinghan Zhang, Tangjun Su, Junteng Liu, Chuancheng Lv, Yikai Zhang, Yao Fu, et~al. 2024{\natexlab{b}}.
\newblock C-eval: A multi-level multi-discipline chinese evaluation suite for foundation models.
\newblock \emph{Advances in Neural Information Processing Systems}, 36.

\bibitem[{iFLYTEK()}]{Spark}
iFLYTEK.
\newblock \href {https://xinghuo.xfyun.cn/} {Spark}.

\bibitem[{Jimenez et~al.(2023)Jimenez, Yang, Wettig, Yao, Pei, Press, and Narasimhan}]{jimenez2023swe}
Carlos~E Jimenez, John Yang, Alexander Wettig, Shunyu Yao, Kexin Pei, Ofir Press, and Karthik Narasimhan. 2023.
\newblock Swe-bench: Can language models resolve real-world github issues?
\newblock \emph{arXiv preprint arXiv:2310.06770}.

\bibitem[{Li et~al.(2023{\natexlab{a}})Li, Zhang, Koto, Yang, Zhao, Gong, Duan, and Baldwin}]{li2023cmmlu}
Haonan Li, Yixuan Zhang, Fajri Koto, Yifei Yang, Hai Zhao, Yeyun Gong, Nan Duan, and Timothy Baldwin. 2023{\natexlab{a}}.
\newblock Cmmlu: Measuring massive multitask language understanding in chinese.
\newblock \emph{arXiv preprint arXiv:2306.09212}.

\bibitem[{Li et~al.(2023{\natexlab{b}})Li, Zhang, Dubois, Taori, Gulrajani, Guestrin, Liang, and Hashimoto}]{alpaca_eval}
Xuechen Li, Tianyi Zhang, Yann Dubois, Rohan Taori, Ishaan Gulrajani, Carlos Guestrin, Percy Liang, and Tatsunori~B. Hashimoto. 2023{\natexlab{b}}.
\newblock Alpacaeval: An automatic evaluator of instruction-following models.
\newblock \url{https://github.com/tatsu-lab/alpaca_eval}.

\bibitem[{Lin et~al.(2024)Lin, Chandu, Brahman, Deng, Ravichander, Pyatkin et~al.}]{wildbench2024}
Bill~Yuchen Lin, Khyathi Chandu, Faeze Brahman, Yuntian Deng, Abhilasha Ravichander, Valentina Pyatkin, et~al. 2024.
\newblock \href {https://huggingface.co/spaces/allenai/WildBench} {Wildbench: Benchmarking llms with challenging tasks from real users in the wild}.

\bibitem[{Liu et~al.(2023{\natexlab{a}})Liu, Lei, Wang, Huang, Feng, Wen, Cheng, Ke, Xu, Tam et~al.}]{liu2023alignbench}
Xiao Liu, Xuanyu Lei, Shengyuan Wang, Yue Huang, Zhuoer Feng, Bosi Wen, Jiale Cheng, Pei Ke, Yifan Xu, Weng~Lam Tam, et~al. 2023{\natexlab{a}}.
\newblock Alignbench: Benchmarking chinese alignment of large language models.
\newblock \emph{arXiv preprint arXiv:2311.18743}.

\bibitem[{Liu et~al.(2023{\natexlab{b}})Liu, Yu, Zhang, Xu, Lei, Lai, Gu, Ding, Men, Yang et~al.}]{liu2023agentbench}
Xiao Liu, Hao Yu, Hanchen Zhang, Yifan Xu, Xuanyu Lei, Hanyu Lai, Yu~Gu, Hangliang Ding, Kaiwen Men, Kejuan Yang, et~al. 2023{\natexlab{b}}.
\newblock Agentbench: Evaluating llms as agents.
\newblock \emph{arXiv preprint arXiv:2308.03688}.

\bibitem[{Liu et~al.(2023{\natexlab{c}})Liu, Iter, Xu, Wang, Xu, and Zhu}]{liu2023gpteval}
Yang Liu, Dan Iter, Yichong Xu, Shuohang Wang, Ruochen Xu, and Chenguang Zhu. 2023{\natexlab{c}}.
\newblock Gpteval: Nlg evaluation using gpt-4 with better human alignment.
\newblock \emph{arXiv preprint arXiv:2303.16634}.

\bibitem[{MoonshotAI()}]{Moonshot}
MoonshotAI.
\newblock \href {https://kimi.moonshot.cn/} {Moonshot}.

\bibitem[{Qin et~al.(2023)Qin, Liang, Ye, Zhu, Yan, Lu, Lin, Cong, Tang, Qian, Zhao, Tian, Xie, Zhou, Gerstein, Li, Liu, and Sun}]{qin2023toolllm}
Yujia Qin, Shihao Liang, Yining Ye, Kunlun Zhu, Lan Yan, Yaxi Lu, Yankai Lin, Xin Cong, Xiangru Tang, Bill Qian, Sihan Zhao, Runchu Tian, Ruobing Xie, Jie Zhou, Mark Gerstein, Dahai Li, Zhiyuan Liu, and Maosong Sun. 2023.
\newblock \href {http://arxiv.org/abs/2307.16789} {Toolllm: Facilitating large language models to master 16000+ real-world apis}.

\bibitem[{Shah et~al.(2023)Shah, White, Andersen, Buscher, Counts, Das, Montazer, Manivannan, Neville, Ni et~al.}]{shah2023using}
Chirag Shah, Ryen~W White, Reid Andersen, Georg Buscher, Scott Counts, Sarkar Snigdha~Sarathi Das, Ali Montazer, Sathish Manivannan, Jennifer Neville, Xiaochuan Ni, et~al. 2023.
\newblock Using large language models to generate, validate, and apply user intent taxonomies.
\newblock \emph{arXiv preprint arXiv:2309.13063}.

\bibitem[{Sottana et~al.(2023)Sottana, Liang, Zou, and Yuan}]{sottana2023evaluation}
Andrea Sottana, Bin Liang, Kai Zou, and Zheng Yuan. 2023.
\newblock Evaluation metrics in the era of gpt-4: reliably evaluating large language models on sequence to sequence tasks.
\newblock \emph{arXiv preprint arXiv:2310.13800}.

\bibitem[{Suzgun et~al.(2022)Suzgun, Scales, Schärli, Gehrmann, Tay, Chung, Chowdhery, Le, Chi, Zhou, and Wei}]{suzgun2022challenging}
Mirac Suzgun, Nathan Scales, Nathanael Schärli, Sebastian Gehrmann, Yi~Tay, Hyung~Won Chung, Aakanksha Chowdhery, Quoc~V. Le, Ed~H. Chi, Denny Zhou, and Jason Wei. 2022.
\newblock \href {http://arxiv.org/abs/2210.09261} {Challenging big-bench tasks and whether chain-of-thought can solve them}.

\bibitem[{Wang et~al.(2023{\natexlab{a}})Wang, Li et~al.}]{wang2023enabling}
Bryan Wang, Gang Li, et~al. 2023{\natexlab{a}}.
\newblock Enabling conversational interaction with mobile ui using large language models.
\newblock In \emph{Proceedings of the 2023 CHI Conference on Human Factors in Computing Systems}.

\bibitem[{Wang et~al.(2023{\natexlab{b}})Wang, Zhong, Li, Mi, Zeng, Huang, Shang, Jiang, and Liu}]{wang2023aligning}
Yufei Wang, Wanjun Zhong, Liangyou Li, Fei Mi, Xingshan Zeng, Wenyong Huang, Lifeng Shang, Xin Jiang, and Qun Liu. 2023{\natexlab{b}}.
\newblock Aligning large language models with human: A survey.
\newblock \emph{arXiv preprint arXiv:2307.12966}.

\bibitem[{Xie et~al.(2023)Xie, Yao, Dai, Wang, Zhou, Jin, Feng, Wei, Lin, Hu et~al.}]{xie2023tencentllmeval}
Shuyi Xie, Wenlin Yao, Yong Dai, Shaobo Wang, Donlin Zhou, Lifeng Jin, Xinhua Feng, Pengzhi Wei, Yujie Lin, Zhichao Hu, et~al. 2023.
\newblock Tencentllmeval: a hierarchical evaluation of real-world capabilities for human-aligned llms.
\newblock \emph{arXiv preprint arXiv:2311.05374}.

\bibitem[{Yang et~al.(2023)Yang, Xiao, Wang, Zhang, Bian, Yin, Lv, Pan, Wang, Yan et~al.}]{Baichuan}
Aiyuan Yang, Bin Xiao, Bingning Wang, Borong Zhang, Ce~Bian, Chao Yin, Chenxu Lv, Da~Pan, Dian Wang, Dong Yan, et~al. 2023.
\newblock Baichuan 2: Open large-scale language models.
\newblock \emph{arXiv preprint arXiv:2309.10305}.

\bibitem[{Zellers et~al.(2019)Zellers, Holtzman, Bisk, Farhadi, and Choi}]{zellers2019hellaswag}
Rowan Zellers, Ari Holtzman, Yonatan Bisk, Ali Farhadi, and Yejin Choi. 2019.
\newblock Hellaswag: Can a machine really finish your sentence?
\newblock \emph{arXiv preprint arXiv:1905.07830}.

\bibitem[{Zhao et~al.(2023)Zhao, Ren, Hessel, Cardie, Choi, and Deng}]{zhao2023inthe}
Wenting Zhao, Xiang Ren, Jack Hessel, Claire Cardie, Yejin Choi, and Yuntian Deng. 2023.
\newblock (inthe) wildchat: 570k chatgpt interaction logs in the wild.
\newblock In \emph{The Twelfth International Conference on Learning Representations}.

\bibitem[{Zheng et~al.(2024)Zheng, Chiang, Sheng, Zhuang, Wu, Zhuang, Lin, Li, Li et~al.}]{zheng2024judging}
Lianmin Zheng, Wei-Lin Chiang, Ying Sheng, Siyuan Zhuang, Zhanghao Wu, Yonghao Zhuang, Zi~Lin, Zhuohan Li, Dacheng Li, et~al. 2024.
\newblock Judging llm-as-a-judge with mt-bench and chatbot arena.
\newblock \emph{Advances in Neural Information Processing Systems}, 36.

\end{thebibliography}
